\PassOptionsToPackage{table}{xcolor}
\PassOptionsToPackage{sort&compress}{natbib}
\documentclass[]{fairmeta}
\setcitestyle{numbers,square,comma}

\usepackage{amsmath,amsfonts,bm}

\def\eqref#1{equation~\ref{#1}}

\def\1{\bm{1}}

\DeclareMathAlphabet{\mathsfit}{\encodingdefault}{\sfdefault}{m}{sl}
\SetMathAlphabet{\mathsfit}{bold}{\encodingdefault}{\sfdefault}{bx}{n}

\usepackage{amsmath}
\usepackage{amssymb}
\usepackage{algorithm}
\usepackage{algpseudocode}
\algrenewcommand\algorithmicrequire{\textbf{Input:}}
\algrenewcommand\algorithmicensure{\textbf{Output:}}
\providecommand{\sg}[1]{\mathrm{sg}[#1]}
\usepackage{array}
\usepackage{booktabs}
\usepackage{tabularx}
\newcolumntype{Y}{>{\raggedleft\arraybackslash}X}
\usepackage{enumitem}
\usepackage{graphicx}
\usepackage{microtype}
\usepackage{multirow}
\usepackage{url}

\definecolor{draftgray}{gray}{0.42}
\definecolor{blue}{RGB}{235, 245, 251}
\newcommand{\method}{FlashForward}
\newcommand{\projectpage}[1][]{\href{https://yikai-wang.github.io/FlashForward/#1}{project page}}

\definecolor{stalemeas}{HTML}{CC0000}

\newcommand{\ms}[2]{\shortstack[r]{#1\\[-0.15em]{\tiny$\pm$#2}}}
\newcommand{\mslab}[1]{\shortstack[l]{#1\\[-0.15em]{\tiny\strut}}}

\title{In-Flight KV Cache with Clean Anchors for Faster Autoregressive Video Diffusion}

\author[1]{Yikai~Wang}
\author[1]{Xiao~Han}
\author[1]{Mengmeng~Xu}
\author[1]{Juan~C.~P\'{e}rez}
\author[1]{Yiannis~Douratsos}
\author[1]{Sen~He}
\author[1]{Zijian~Zhou}
\author[1]{Fei~Zhang}
\author[1]{Zhaochong~An}
\author[1]{Juan-Manuel~P\'{e}rez-R\'{u}a}
\author[2]{Chen~Change~Loy}
\author[1]{Tao~Xiang}

\affiliation[1]{Meta}
\affiliation[2]{S-Lab, Nanyang Technological University}

\abstract{
Few-step autoregressive video diffusion generates a long video by splitting the video into temporal chunks and generating chunk-by-chunk, each through a short sequence of denoising stages.
To memorize chunks that are already generated, previous methods reconstruct a clean or less-noisy key--value (KV) cache by additional forwards to build the cache without advancing an output latent.
However, every denoising forward itself already computes the in-flight KV of the current chunk.
We introduce \method{}, which directly reuses this cache to avoid the heavy cache-update-only model forwards.
After the current chunk completes one denoising stage, its stage-specific cache is already available for the next chunk.
Assigning one GPU to each stage therefore lets different chunks occupy different stages concurrently.
This early availability has a quality cost: the resulting stage-matched history is noisy, causing appearance and motion drift among chunks.
To complement it, \method{} produces sparse auxiliary clean anchor latents before the corresponding region is generated so the generation trajectories can be stabilized by this two-sided conditioning.
The two memories operate at different temporal scales: sparse clean anchor KV supplies coarse, long-range two-sided structural guidance, while dense stage-matched history preserves fine, recent evolution.
With up to four GPUs, \method{} runs $1.16$--$1.69\times$ faster than HiAR and $1.42$--$2.92\times$ faster than Self-Forcing for 16 FPS videos of 20 seconds or longer across 1.3B and 14B backbone scales at 480p and 720p.
On VBench, for the 1.3B model at 480p, it achieves higher scores and remains stable at longer durations, demonstrating that \method{} generates high-quality and temporally consistent videos across durations of 20s, 35s and 65s at a much faster generation speed.
}

\date{\today}
\metadata[Project Page]{\url{https://yikai-wang.github.io/FlashForward}}

\begin{document}

\maketitle

\begin{figure}[h!]
\centering
\includegraphics[width=\linewidth]{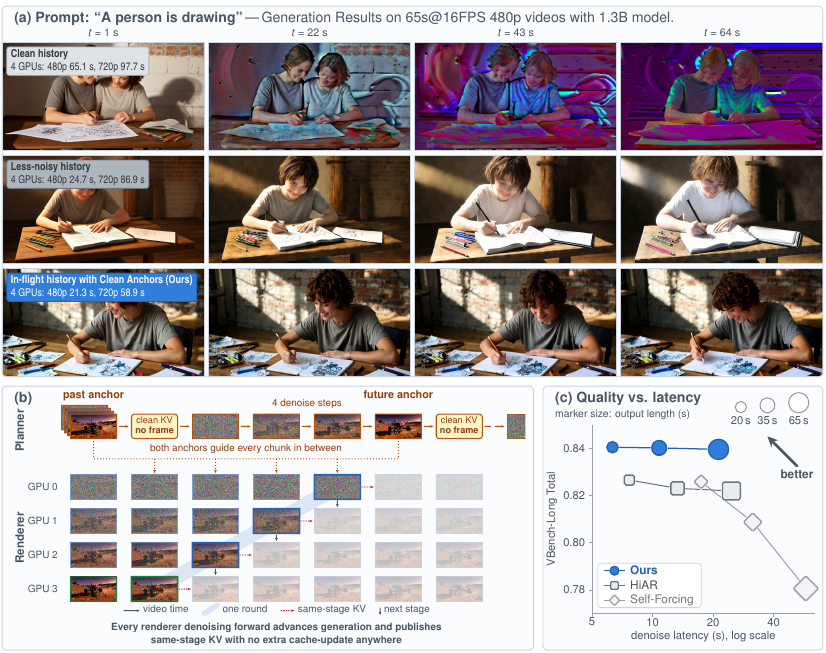}
\caption{
\textbf{
\method{} redesigns the autoregressive diffusion generation pipeline to remove heavy cache-update-only forwards, significantly reducing generation latency}.
\textbf{(a)} Compared with pipelines that  reconstruct clean (Self-Forcing~\citep{huang2025selfforcing}) or less-noisy (HiAR~\citep{zou2026hiar}) KV through context encoding that does not advance an output latent, \method{} substantially reduces latency while preserving generation quality and coherence, even in long videos.
\textbf{(b)} \method{} adopts a planner to autoregressively produce sparse auxiliary anchors for two-sided temporal conditioning.
A renderer can then run in standard autoregressive form conditioned on these anchors such that every ordinary forward advances one chunk and publishes stage-matched history for later chunks, allowing different denoising stages to run on different GPUs.
\textbf{(c)} Quality versus latency for 1.3B models on 480p videos, using each method's fastest setup with up to four GPUs.
\emph{See the \projectpage[\#comparison]{} for an interactive demo.}
}
\label{fig:state-contract}
\end{figure}

\section{Introduction}
Autoregressive video diffusion~\citep{ho2022videodiffusion,voleti2022mcvd,harvey2022flexible} generates a video one short chunk at a time.
Each chunk is produced by generating its latent through a sequence of denoising stages.
To memorize generated content, the model stores features from earlier chunks in a key--value (KV) cache for subsequent generation with additional model passes~\citep{gao2025ca2vdm,deng2025nova,sandai2025magi1}.
Recent methods reduce the denoising sequence to only a few stages~\citep{yin2025causvid,huang2025selfforcing,zheng2026causalrcm,liu2026rollingforcing,liu2026diagonaldistillation}, making any additional model forward used only to prepare memory an increasingly important bottleneck.

Existing methods incur this overhead in different ways.
Self-Forcing~\citep{huang2025selfforcing} completes a chunk and then passes its clean output through the model again to construct the cache.
HiAR~\citep{zou2026hiar} allows different segments to overlap in execution by conditioning on less-noisy rather than clean-endpoint context, but re-encoding at every denoising stage.
These additional passes build memory without directly advancing video generation.
However, the computation during denoising already establishes an \emph{in-flight KV} for the current chunk.
If we directly reuse this cache, we could avoid the heavy cache-update-only model forwards and further reduce the generation latency.

More broadly, we analyze the design choice of the \emph{cross-chunk state contract}, specifying what representation each chunk publishes, at what noise level, and when later chunks may consume it.
The contract determines not only the temporal memory available to the model, but also the dependency graph of generation.
In previous clean or less-noisy history contracts, the cache-update-only forwards build state without advancing an output latent.
In this paper, we propose \method{} to instead publish an ordinary same-stage cache for reuse at the same denoising stage, making the required state available as part of generation itself.
Among the designs in Tab.~\ref{tab:state-contract}, it is the only memory that is both emitted by an output-advancing forward and available early to pipeline chunks across stage workers.

This early availability has a cost.
Stage-matched history comes from unfinished, noisy chunks.
Used alone, it can propagate uncertainty in appearance and motion~\citep{chen2024diffusionforcing,zou2026hiar}.
Returning to dense clean state for every completed chunk would restore the heavy cache-update-only work.
\method{} therefore complements this promptly available history with sparse auxiliary anchors distributed across the timeline and planned in advance for a clean two-sided memory.

This availability rule produces a planner--renderer generation graph.
The \emph{planner role} produces the auxiliary anchor latents and their clean anchor KV cache bank; the \emph{renderer role} generates the video while reading a local anchor KV block and the stage-matched renderer history bank, as in Fig.~\ref{fig:state-contract}(b).
The two memories divide temporal responsibility: sparse clean anchor KV constrains coarse structure over a longer interval, whereas dense stage-matched renderer history carries fine appearance and motion changes from recent chunks.
Only sparse planner blocks require a clean-endpoint cache-extraction forward, and each local anchor window conditions multiple renderer chunks.
Hence the efficiency of dense rendering without cache-update-only forwards is preserved.

\method{} uses a shared generator backbone for both roles.
A learned role embedding and role-specific LoRA adapters~\citep{hu2022lora} specialize their temporal dependencies while retaining common visual and language knowledge.
We train \method{} in two phases:
Packed supervised fine-tuning establishes the planner--renderer graph from real videos; self-rollout distillation distills the guidance and denoising stages, and adapts its few-step generation process to generated context.

We evaluate and compare the latency of \method{} with other generation pipelines across 1.3B and 14B backbone scales on 480p and 720p videos.
For 16 FPS videos of 20 seconds or longer in up to four-GPU settings,
\method{} is $1.16$--$1.69\times$ faster than less-noisy history (HiAR) and $1.42$--$2.92\times$ faster than clean history (Self-Forcing).
We evaluate generation quality for the 1.3B model at 480p under VBench.
\method{} scores 0.838 on Total, surpassing 0.805 for Self-Forcing and 0.821 for HiAR.
Furthermore, as the duration increases to 65 seconds, the generation quality remains stable for \method{}, demonstrating its effectiveness as a faster generation pipeline.

\paragraph{Contributions.}
Our contribution is one state-availability design, realized as a single causal chain.
\textbf{(1)} Each ordinary renderer forward publishes stage-matched renderer history while advancing the current chunk, providing state available early enough to induce an inter-chunk wavefront over stage workers without renderer cache-update-only forwards.
\textbf{(2)} We instantiate a temporal-scale allocation of cross-chunk state: sparse clean anchor KV prepared ahead of each local region supplies coarse long-range structure, while promptly available but noisy and past-only renderer history supplies fine recent evolution.
\textbf{(3)} We realize this graph with planner and renderer roles sharing a base generator, train it through real-video supervision and self-rollout distillation, and demonstrate latency benefits across model sizes and resolutions together with quality superiority for the 1.3B generator at 480p.

\section{Related Work}

\begin{table}[t]
\caption{\textbf{How memory determines execution} for $S$ denoising stages with noise levels $\sigma_S>\cdots>\sigma_0=0$.
\method{} additionally uses sparse clean anchor KV prepared before generation.}
\label{tab:state-contract}
\scriptsize
\center{
\begin{tabular*}{\linewidth}{@{\extracolsep{\fill}}lclll@{}}
\toprule
Method & Memory noise level & When it becomes usable & Cache-update-only forwards & Chunk execution \\
\midrule
Self-Forcing
& $\sigma_0$
& After the chunk is re-encoded
& $1$ per chunk
& Serial \\
N-C-Causal-rCM
& $\sigma_1\rightarrow\sigma_0$
& After the last denoising stage
& $0$
& Serial \\
HiAR
& $\sigma_{k-1}$
& After each stage-specific re-encoding
& $S$ per chunk
& Overlapped \\
\textbf{\method{}}
& $\sigma_k\rightarrow\sigma_{k-1}$
& Immediately after each denoising stage
& $0$
& Overlapped \\
\bottomrule
\end{tabular*}
}
\end{table}

\paragraph{Few-step autoregressive video diffusion.}
Video generators combine diffusion or flow objectives with pixel-space, latent, and Transformer backbones~\citep{ho2020ddpm,ho2022videodiffusion,lipman2023flowmatching,singer2023makeavideo,rombach2022latent,blattmann2023align,blattmann2023stablevideo,peebles2023dit}.
Few-step objectives use implicit sampling, distillation, and consistency training~\citep{song2021ddim,song2023consistency,salimans2022progressive,zhai2024motionconsistency,cai2026modemeanseeking}.
Autoregressive video models factor generation into sequential units and generated histories~\citep{weissenborn2020scaling,villegas2022phenaki,weng2024artv,henschel2025streamingt2v}; \method{} changes what is passed between those units, and with it their execution order.

\paragraph{Cross-chunk memory and execution.}
We view an autoregressive diffusion method as defining a \emph{cross-chunk state contract}: which representation a chunk publishes, at what noise level, and when that representation becomes available to later chunks.
Together with the within-chunk denoising order, this contract determines both the number of cache-update-only forwards and which chunk updates can overlap (Tab.~\ref{tab:state-contract}).
Self-Forcing~\citep{huang2025selfforcing} publishes clean KV only after a completed chunk undergoes one cache-update-only forward, so later chunks wait at a serial boundary.
The noisy-context variant of Causal-rCM~\citep{zheng2026causalrcm} (N-C-Causal-rCM), following Diagonal Distillation~\citep{liu2026diagonaldistillation}, reuses KV from the final denoising forward, avoiding that extra update; because the state is published only when the predecessor reaches its final stage, chunks remain serial.
HiAR~\citep{zou2026hiar} conditions each chunk at intermediate stage on its predecessors' less-noisy context, enabling an anti-diagonal pipeline at the cost of per-stage context encoding per consumed chunk.
\method{} instead publishes the KV produced by every ordinary renderer forward to later chunks at the same stage.
Each publication advances the current chunk rather than serving only to build memory, and the resulting same-stage dependencies form a wavefront.
Because this stage-matched renderer history is noisy and past-only, \method{} complements it with sparse clean anchor KV made available before each local region is rendered.
In audio-driven talking-avatar generation, TalkingMachines~\citep{low2025talkingmachines} and Live Avatar~\citep{huang2025liveavatar} condition on stage-matched history, and Live Avatar also pipelines denoising timesteps across GPUs.
Their success relies on a given clean reference image that anchors appearance and on a largely static scene.
Text-to-video provides neither; there, stage-matched history alone degrades quality in HiAR~\citep{zou2026hiar} and fails in our ablation (Tab.~\ref{tab:conditioning-modules}).
The clean anchor KV of \method{} is what makes stage-matched history usable in text-to-video, generalizing the single given reference image of talking-avatar generation to sparse planner-generated anchors throughout the video.
Another research direction~\citep{gu2025far,yang2026longlive,samuel2026fastar,ji2026forcingkv,ma2026flowcache,luo2026futureforcing,yang2026anchorforcing} studies how to determine which cached content to retain, compress, merge, reuse, or switch; these policies are largely orthogonal to the state contract.

\paragraph{Plan-then-infill and future-guided generation.}
Sparse-to-dense video generation first establishes coarse temporal structure and then fills the intervening frames~\citep{ge2022tats,he2022lvdm,harvey2022flexible,yin2023nuwaxl}.
\citet{xiang2025mmpl} and~\citet{ouyang2026dcarl} separate keyframe planning from segment population; FramePack~\citep{zhang2025framepack} and SneakPeek~\citep{hong2025sneakpeek} predict anchors and future keyframes ahead of the frames between them;~\citet{bendel2026ats} and~\citet{zhang2026unitemp} use two-sided anchors for interval generation.
These works establish sparse planning and two-sided conditioning as effective sources of temporal coherence.
\method{} repurposes this form of conditioning as a coarse-timescale state that complements fine-timescale stage-matched renderer history.
Unlike in conventional infilling, its auxiliary anchor latents are conditioning-only.
Their clean anchor KV lets the renderer retain promptly available stage-matched renderer history without reverting to dense clean cache-update-only forwards.

\paragraph{Noise schedules and parallel execution.}
Diffusion Forcing~\citep{chen2024diffusionforcing}, FIFO-Diffusion~\citep{kim2024fifodiffusion}, Rolling Forcing~\citep{liu2026rollingforcing}, and Diagonal Distillation~\citep{liu2026diagonaldistillation} assign different noise levels or update times across temporal positions, while Ms.\ Forcing~\citep{li2026msforcing} adapts computation to the noise level.
DistriFusion~\citep{li2024distrifusion} and PipeFusion~\citep{fang2025pipefusion} parallelize spatial computation within a denoising trajectory.
HiAR~\citep{zou2026hiar} pipelines successive chunks across denoising-stage workers, placing them on an anti-diagonal schedule that it sustains by re-encoding each consumed predecessor at every stage.
Its context-only work is therefore paid per stage per consumed chunk, rather than once per chunk as in Self-Forcing.
A noise schedule specifies when each chunk is updated, whereas a cross-chunk state rule specifies which earlier representation conditions that update.
In \method{}, the same anti-diagonal follows directly from the same-stage state dependency, and no output chunk is re-encoded.

\section{Stage-Matched History with Clean Two-Sided Anchors}

\method{} uses one generator in two roles.
The planner first produces a sparse clean plan, and the renderer then generates the video while reading that plan and the recent history left by earlier renderer chunks.
The key operation is one ordinary renderer forward: it both advances the current chunk and leaves reusable KV for later renderer chunks during generation.

\begin{figure}[t]
\centering
\includegraphics[width=\linewidth]{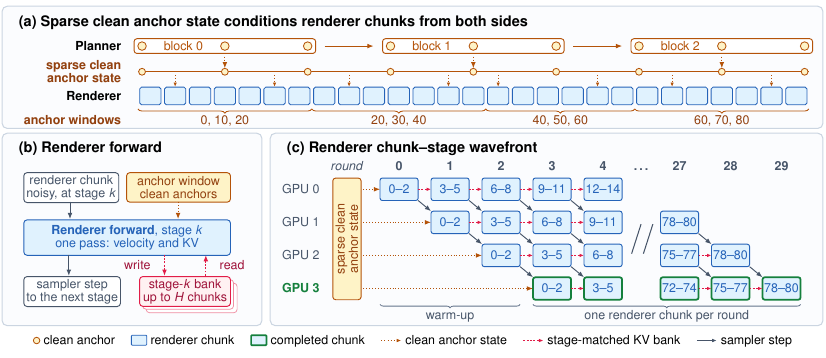}
\caption{
\textbf{(a)} The planner produces sparse auxiliary anchor latents; local three-anchor windows guide the renderer.
\textbf{(b)} One renderer forward reads clean anchor KV and stage-matched renderer history, predicts its latent update, and publishes its own KV.
\textbf{(c)} These dependencies form a wavefront that completes one chunk per pipeline round after warm-up.
See the concrete generation example in Sec.~\ref{sec:example}.
\emph{See the \projectpage[\#generation]{} for an interactive demo.}
}
\label{fig:generation}
\end{figure}

\paragraph{A concrete generation example.}
\label{sec:example}
Fig.~\ref{fig:generation} follows an output of 81 latent frames.
The planner generates nine auxiliary clean anchor latents at positions $\{0,10,\ldots,80\}$, three anchors at a time.
It runs one cache-extraction forward on each completed block to obtain clean anchor KV.
The renderer then generates all 81 output positions in 27 three-latent chunks guided by a two-sided clean plan and in-flight memory.
For example, the chunk at positions 6--8 reads nearby clean anchor KV at $\{0,10,20\}$, which bracket this interior chunk, together with the KV left by recent chunks at its current denoising stage.
As each chunk traverses four stages, the next chunk can follow one stage behind, producing the wavefront in panel (c).
The complete data flow is therefore: generate each auxiliary anchor block and publish its clean KV, then render the dense video while every renderer forward advances the output and publishes stage-matched renderer history.

\paragraph{Notation.}
Given a condition $c$, the target contains $L$ latent positions indexed by $\mathcal{I}=\{0,\ldots,L-1\}$.
The \emph{auxiliary anchor} indices form a sparse subset of this timeline with anchor stride $\Delta$ as
\begin{equation}
\mathcal{P}=\{m\Delta\mid m\in\mathbb{N}_0,\ m\Delta<L\}\cup\{L-1\},
\qquad
M=|\mathcal{P}|.
\label{eq:anchor-infill-partition}
\end{equation}
The planner processes $\mathcal{P}$ in $N_{\mathrm P}=\lceil M/B_{\mathrm P}\rceil$ ordered blocks $\{\mathcal{Q}_j\}$ of at most $B_{\mathrm P}$ auxiliary anchor latents.
The renderer processes $\mathcal{I}$ in $N_{\mathrm R}=\lceil L/B_{\mathrm R}\rceil$ ordered chunks $\{\mathcal{C}_i\}$ of at most $B_{\mathrm R}$ latents, and $\Gamma(i)=\{2u\Delta,(2u{+}1)\Delta,(2u{+}2)\Delta\}\subseteq\mathcal{P},u=\lfloor B_{\mathrm R}i/(2\Delta)\rfloor$ identifies the clean anchor KV entries visible to renderer chunk $i$.
Interior regions use anchor windows with indices before and after each region; at the beginning or end of a finite video, $\Gamma(i)$ uses the available boundary-side anchors.
Let $\theta$ denote the shared model with role-specific embeddings and LoRA adapter sets, while $F_{\theta}^{\mathrm P}$ and $F_{\theta}^{\mathrm R}$ denote its planner and renderer routes.
At inference, both use $S$ denoising stages indexed by $k$, $\sigma_S>\cdots>\sigma_0=0$, and $\tau_k=\tau(\sigma_k)$ denotes the corresponding timestep.
Our empirical configuration uses $B_{\mathrm P}=B_{\mathrm R}=3$, $\Delta=10$, $S=4$, a context budget of $W_{\mathrm{ctx}}=21$ latents, a renderer history length of $H=5$ chunks, and a planner history length of $H_{\mathrm P}=6$ blocks.
Sec.~\ref{app:anchor-rate} explains how equal planner block and renderer chunk sizes together with $\Delta=10$ balance anchor production and consumption during streaming generation.

\subsection{Stage-Matched Renderer History Enables Pipelined Rendering}

\label{sec:temporal-state}
\label{sec:wavefront}

A renderer forward takes the current chunk's noisy latents at a given denoising stage as \emph{inputs}.
It reads KV from earlier chunks at the same stage together with clean KV from nearby anchors as \emph{memory}.
The forward produces the prediction used to advance denoising as \emph{outputs} and stores its own KV for later chunks at that stage as \emph{updated memory}.
See Fig.~\ref{fig:generation}(b) for an illustration.

Precisely, $\mathcal{A}_{\Gamma(i)}$ contains clean anchor KV entries selected by $\Gamma(i)$.
The stage-matched renderer history bank $\mathcal{B}_{k}^{(i)}$ contains KV produced when up to $H$ preceding renderer chunks passed through denoising stage $k$.
Under the rectified-flow parameterization~\citep{liu2023rectifiedflow}, $x_\sigma=(1-\sigma)x_0+\sigma\varepsilon$ and the velocity target is $v=\varepsilon-x_0$.
At stage $k$, the renderer takes the current noisy chunk $x^{\mathrm R}_{i,k}$, conditions on $\mathcal{A}_{\Gamma(i)}$ and $\mathcal{B}_{k}^{(i)}$, and predicts $\widehat v^{\mathrm R}_{i,k}$ to advance the latent from $\sigma_k$ toward $\sigma_{k-1}$.
The same forward publishes its KV to the stage-$k$ bank for later chunks, as shown in Fig.~\ref{fig:generation}(b):
\begin{equation}
\left(\widehat v^{\mathrm R}_{i,k},K^{\mathrm R}_{i,k},V^{\mathrm R}_{i,k}\right)
=F_{\theta}^{\mathrm R}\!\left(x^{\mathrm R}_{i,k},c,\tau_k;
\mathcal{A}_{\Gamma(i)},\mathcal{B}_{k}^{(i)}\right),\
\mathcal{B}_{k}^{(i+1)}
=\operatorname{Tail}_{H}\!\left(
\mathcal{B}_{k}^{(i)}\cup
(i,K^{\mathrm R}_{i,k},V^{\mathrm R}_{i,k})\right).
\label{eq:step-update}
\end{equation}
Both KV banks are ordered sequences: $\cup$ adds their newest entry, and $\operatorname{Tail}_{H}$ retains the $H$ most recent renderer chunks.
No renderer cache-update-only forward.
In practice, the renderer can process the clip autoregressively, updating the KV cache on the fly as it handles one chunk at a time, or process the full clip at once using a chunk-wise causal mask.

\paragraph{Why the two states are complementary.}
The two states are deliberately assigned different temporal scales.
Dense stage-matched renderer history is available immediately and records fine changes in recent appearance and motion, but it is noisy; used alone, it can propagate uncertainty across chunk boundaries.
The sparse clean anchor provides a stable, coarse structural reference over a longer interval for both the past and the future.
The clean-endpoint cache extraction is confined to sparse planner blocks, which advance by a large stride: one planner transition advances 30 latents, compared with 3 for one renderer transition, nearly a tenfold reduction in autoregressive depth.
This reduces the number of per-chunk cache-update-only forwards and leads to inter-chunk pipelining for the renderer.
This coarse plan also improves temporal coherence over long horizons.

\paragraph{Planning the clean anchor KV bank.}
The planner generates auxiliary anchor latents autoregressively across blocks $\{\mathcal{Q}_j\}$.
After each planner block is generated, one planner cache-extraction forward publishes its clean anchor KV entries.
Renderer chunk $i$ reads the local window $\mathcal{A}_{\Gamma(i)}$ with earlier and later anchor indices while adjacent regions share anchors.
In Fig.~\ref{fig:generation}(a), the windows are $\{0,10,20\}$, $\{20,30,40\}$, $\{40,50,60\}$, and $\{60,70,80\}$.

\paragraph{Wavefront rendering.}
A renderer node $(i,k)$ depends on the noisier chunk at node $(i,k{+}1)$ and the earlier chunk at node $(i{-}1,k)$.
With one worker assigned to each of the $S$ stages, nodes on the same anti-diagonal can therefore run concurrently for different chunks, as shown in Fig.~\ref{fig:generation}(c).
A \emph{pipeline round} is one such forward slot on every active stage worker.
After warm-up, the wavefront completes one renderer chunk per round while every forward both denoises and publishes memory.
This pipeline avoids the repeated stage-specific re-encoding in HiAR.
Alg.~\ref{alg:wavefront} gives the batched four-device rollout; device allocation and timing details are deferred to Sec.~\ref{app:latency-results}.

\paragraph{Streaming generation.}
\method{} naturally supports streaming generation by dedicating one
GPU to the planner and the remaining GPUs to the renderer. This allows rendering to begin as soon
as the planner produces the first anchor block, further reducing latency. We discuss this in Sec.~\ref{app:streaming}.

\subsection{Training a Shared Planner and Renderer}
\begin{figure}[t]
\centering
\includegraphics[width=\linewidth]{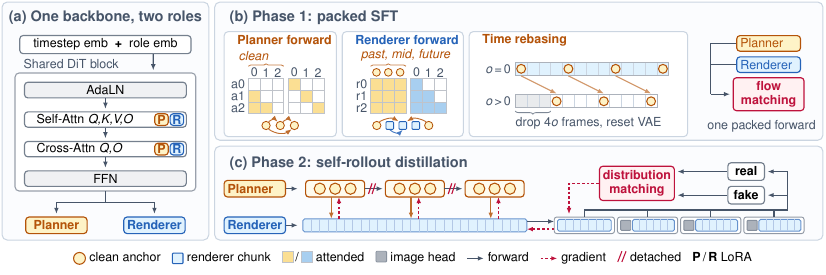}
\caption{
\textbf{(a)} Shared backbone with role-specific embeddings and LoRA adapters.
\textbf{(b)} Packed supervised fine-tuning (SFT) trains both roles in one forward.
\textbf{(c)} Self-rollout distillation adapts the model to long generated contexts.
\emph{See the \projectpage[\#training]{} for an interactive demo.}
}
\vspace{-0.2cm}
\label{fig:training}
\end{figure}

The two roles share visual and language knowledge but require different temporal dependencies.
We use role-specific embeddings and LoRA adapters to specialize these dependencies, as in Fig.~\ref{fig:training}(a).
We learn this in two phases: supervised fine-tuning first establishes both roles from real videos, and self-rollout distillation then adapts them to generated context and guidance-free few-step inference.

\paragraph{Phase 1: Supervised fine-tuning.}
The bidirectional generator is not yet adapted to either the planner's large-stride block-causal dependence or the renderer's clean anchor KV and stage-matched renderer history dependencies.
Phase 1 trains both roles together in one packed forward from real videos.
The planner blocks attend causally to clean KV from earlier blocks as in teacher forcing~\citep{williams1989teacherforcing}.
The renderer targets attend to clean anchor KV admitted by $\Gamma(i)$, and to up to $H$ preceding renderer chunks at the same noise level.
A training clip supplies planner and renderer targets directly.
As planner latents are sparse, we time-rebase each clip by cropping it from temporal offsets for augmentation, as in Fig.~\ref{fig:training}(b).
Alg.~\ref{alg:sft} and Sec.~\ref{app:supervised-details} give the  details.

\paragraph{Phase 2: Self-rollout distillation.}
Phase 1 conditions on ground-truth latents, whereas inference uses planner-generated anchor KV and renderer-generated history throughout a few-step trajectory.
Phase 2 therefore trains on a planner--renderer self-rollout and applies the distribution-matching distillation objective~\citep{yin2024dmd,yin2024dmd2}, as shown in Fig.~\ref{fig:training}.
As in Self-Forcing~\citep{huang2025selfforcing}, conditioning on self-generated context mitigates exposure bias.
The student first performs a causal planner rollout over auxiliary anchor blocks and then  renders the full clip at once.
Both roles follow the same four-stage guidance-free student schedule.
Although the score networks use shorter windows, the loss covers the full clip. 
We split the clip into score-window-sized tiles and decode and re-encode the first frame of each later segment to obtain its image head.
During training, the renderer part is implemented as a standard block-causal forward, rather than a chunk-by-chunk cache update.
This computation graph lets gradients flow from later chunks to earlier chunks, and from the renderer to the planner.
Alg.~\ref{alg:dmd} and Sec.~\ref{app:distillation-details} introduce these in detail.

\section{Experiments}

We build \method{} on Wan2.1-T2V-1.3B~\citep{wan2025} and train it on 256K Shutterstock videos~\citep{shutterstock} with generated captions.
Each clip has 81 latents, corresponding to 321 RGB frames at 16 FPS.
Training takes 1,800 steps for the SFT phase and 100 student steps for the distillation phase with a batch size of 128.
See Sec.~\ref{app:training-details} for details.
We evaluate generation quality on the 1.3B model at $480$p, following previous works.
We use VBench-1.0~\citep{huang2024vbench} on 20-second videos following HiAR~\citep{zou2026hiar}, and VBench-Long~\citep{huang2026vbenchpp} on 20-, 35-, and 65-second videos.
We mainly compare with clean-history methods Self-Forcing~\citep{huang2025selfforcing} and Causal Forcing~\citep{zhu2026causalforcing}, and less-noisy-history method HiAR~\citep{zou2026hiar}.
We also report results from LTX-Video~\citep{hacohen2024ltxvideo}, Wan2.1~\citep{wan2025}, NOVA~\citep{deng2025nova}, Pyramid Flow~\citep{jin2025pyramidalflow}, SkyReels-V2~\citep{chen2025skyreelsv2}, MAGI-1~\citep{sandai2025magi1}, and CausVid~\citep{yin2025causvid} for reference.

\begin{figure}[t]
\centering
\includegraphics[width=\linewidth]{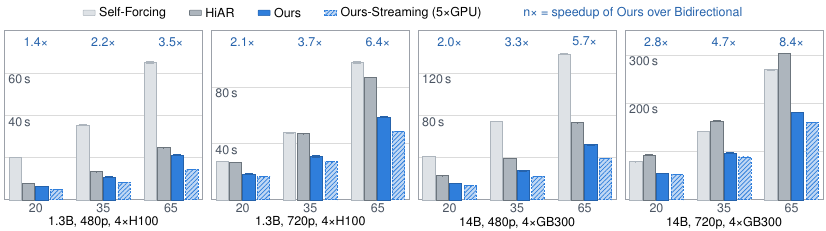}
\caption{
Average denoising latency on four-GPUs and our five-GPU streaming version with standard deviations. Tab.~\ref{tab:generation-time} reports exact numbers.
\emph{See the \projectpage[\#latency]{} for an interactive demo.}
}
\label{fig:latency}
\end{figure}
\begin{table}[t]
\caption{VBench-1.0 on 20\,s videos (top; 5\,s for bidirectional models) and VBench-Long (bottom).
}
\label{tab:vbench}
\label{tab:vbench-long}
\begin{center}
\scriptsize
\setlength{\tabcolsep}{1pt}
\begin{tabular*}{\linewidth}{@{\extracolsep{\fill}}lrrrrrrrrrrr@{}}
\toprule
\textit{VBench-1.0} & LTX- & Wan2.1- & \multirow{2}{*}{NOVA} & Pyramid & SkyReels- & MAGI-1-
  & \multirow{2}{*}{CausVid} & Self- & Causal
  & \multirow{2}{*}{HiAR} & \multirow{2}{*}{\textbf{\method{}}} \\
Metric & Video & T2V-1.3B & & Flow & V2-1.3B & 4.5B &
  & Forcing & Forcing & & \\
\midrule
Total $\uparrow$
  & 0.766 & 0.802 & 0.773 & 0.775 & 0.788 & 0.757 & 0.764
  & 0.805 & 0.810 & 0.821 & \textbf{0.838} \\
Quality $\uparrow$
  & 0.789 & 0.813 & 0.777 & 0.804 & 0.808 & 0.785 & 0.771
  & 0.829 & 0.837 & 0.846 & \textbf{0.859} \\
Semantic $\uparrow$
  & 0.685 & \textbf{0.766} & 0.757 & 0.670 & 0.707 & 0.647 & 0.740
  & 0.708 & 0.701 & 0.723 & 0.753 \\
\bottomrule
\end{tabular*}

\scriptsize
\setlength{\tabcolsep}{1pt}
\begin{tabular*}{\linewidth}{@{\extracolsep{\fill}}lcccccccccccc@{}}
\toprule
\textit{VBench-Long}
  & \multicolumn{3}{c}{Self-Forcing}
  & \multicolumn{3}{c}{Causal Forcing}
  & \multicolumn{3}{c}{HiAR}
  & \multicolumn{3}{c}{\textbf{\method{}}} \\
\cmidrule(lr){2-4}\cmidrule(lr){5-7}\cmidrule(lr){8-10}\cmidrule(lr){11-13}
Duration  & Total $\uparrow$ & Quality $\uparrow$ & Semantic $\uparrow$
  & Total $\uparrow$ & Quality $\uparrow$ & Semantic $\uparrow$
  & Total $\uparrow$ & Quality $\uparrow$ & Semantic $\uparrow$
  & Total $\uparrow$ & Quality $\uparrow$ & Semantic $\uparrow$ \\
\midrule
20\,s
  & 0.8258 & 0.8460 & 0.7452
  & 0.8178 & 0.8515 & 0.6833
  & 0.8265 & 0.8440 & \textbf{0.7565}
  & \textbf{0.8404} & \textbf{0.8635} & 0.7481 \\
35\,s
  & 0.8087 & 0.8330 & 0.7116
  & 0.7906 & 0.8343 & 0.6159
  & 0.8230 & 0.8452 & 0.7339
  & \textbf{0.8401} & \textbf{0.8636} & \textbf{0.7460} \\
65\,s
  & 0.7806 & 0.8186 & 0.6288
  & 0.7539 & 0.8145 & 0.5113
  & 0.8219 & 0.8419 & 0.7418
  & \textbf{0.8395} & \textbf{0.8617} & \textbf{0.7505} \\
\bottomrule
\end{tabular*}
\end{center}
\end{table}

\subsection{Fast and High-Quality Generation}
\label{sec:efficiency}
\label{sec:quality}

\paragraph{Faster generation.}
Fig.~\ref{fig:latency} and Tab.~\ref{tab:generation-time} show that, from 20\,s onward and using each method's fastest setup with up to four GPUs, \method{} is the fastest autoregressive schedule in every model--resolution setting, $1.16$--$1.69\times$ faster than HiAR and $1.42$--$2.92\times$ faster than Self-Forcing.
The gain comes from two properties of its state contract: stage-matched renderer history avoids the cache-update-only forwards required by Self-Forcing and HiAR, while its availability leads to an inter-chunk stage pipeline.
Sec.~\ref{app:latency-results} provides the detailed analysis.

\begin{figure}[!t]
\centering
\includegraphics[width=\linewidth]{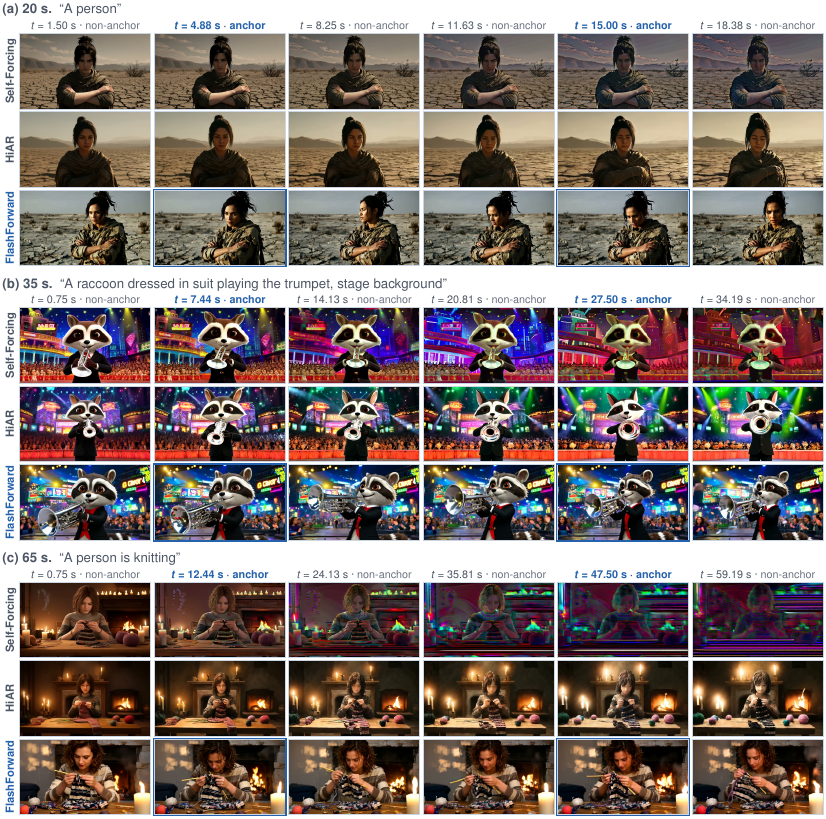}
\caption{
Qualitative comparison at 20, 35, and 65\,s.
Blue boxes indicate anchor frames.
The baselines have no anchors.
\emph{See full prompt and video comparisons on the \projectpage[\#comparison]{}.}
}
\vspace{-0.5cm}
\label{fig:qualitative}
\end{figure}

\paragraph{Quality.}
\textbf{VBench-1.0} results are in Tab.~\ref{tab:vbench} (top).
Competing methods' results are taken from HiAR~\citep{zou2026hiar}.
\method{} achieves the highest Total and Quality scores.
Its Semantic score is the highest among the distilled autoregressive methods.
\textbf{VBench-Long} evaluates long-video quality.
We run all four methods under the same protocol (Tab.~\ref{tab:vbench-long}, bottom).
\method{} leads in Total and Quality at every duration, maintaining long-duration quality, while its semantic alignment remains competitive.
The behavior of \method{} under different durations is stable and consistent.

\paragraph{Qualitative comparison.}
Fig.~\ref{fig:qualitative} shows the same trend qualitatively.
Self-Forcing accumulates color drift.
HiAR avoids this collapse but drifts in other ways: the raccoon's face and ears change in (b) and the knitter's face becomes doll-like in (c), while in (a) the person's face changes and the arms stay blurred under the cloak.
\method{} maintains temporal coherence while preserving motion.
Its frames at anchor positions show no visible difference from those between anchors.

\subsection{Further Analysis}
\label{sec:continuity}

\paragraph{Why regenerate the anchor positions?}
Keeping the planner's anchors as output would be cheaper, but we empirically find that this produces periodic seams at the anchor positions (\emph{see the \projectpage[\#issues]{} for video examples}).
We compare and analyze two variants that both use the planner's anchor latents as part of the video output and differ in the source of the anchor KV: one directly reuses the planner's anchor KV, while the other feeds the anchor latents to the renderer role to obtain renderer-side KV.
Both suffer from a similar seam issue.
We analyze this in details in Sec.~\ref{app:regenerate-anchors}.

\paragraph{Ablation study.}
Tab.~\ref{tab:ablation} (top) reports the training ablations.
Time-rebased supervision provides a better initialization for distillation.
Role specialization lets the model learn role-specific behavior more freely, and
more importantly, reduces seams between anchor and non-anchor positions.

\begin{table}[t]
\caption{Top: training ablations.
Bottom: conditioning-state ablations.
}
\label{tab:ablation}
\label{tab:conditioning-modules}
\begin{center}
\scriptsize
\setlength{\tabcolsep}{2pt}
\begin{tabular*}{\linewidth}{@{\extracolsep{\fill}}lrrrrrr@{}}
\toprule

  & \multicolumn{3}{c}{SFT phase (50-step sampling)}
  & \multicolumn{3}{c}{Distillation phase (4-step sampling)} \\
\cmidrule(lr){2-4}\cmidrule(lr){5-7}

Ablation  & Total $\uparrow$
  & Quality $\uparrow$
  & Semantic $\uparrow$
  & Total $\uparrow$
  & Quality $\uparrow$
  & Semantic $\uparrow$ \\
\midrule
Dual-role SFT
  & 0.8172 & 0.8371 & 0.7376
  & 0.8184 & 0.8332 & 0.7593 \\
$+$\hspace{0.35em}time-rebased supervision
  & 0.8150 & 0.8345 & 0.7371
  & 0.8198 & 0.8338 & \textbf{0.7638} \\
\hspace{0.35em}$+$\hspace{0.35em}role-specific embedding and LoRA
  & \textbf{0.8205} & \textbf{0.8377} & \textbf{0.7516}
  & \textbf{0.8380} & \textbf{0.8593} & 0.7528 \\
\bottomrule
\end{tabular*}
\setlength{\tabcolsep}{3pt}
\begin{tabular*}{\linewidth}{@{\extracolsep{\fill}}lrrr@{}}
\toprule
Conditioning module
  & Total $\uparrow$
  & Quality $\uparrow$
  & Semantic $\uparrow$ \\
\midrule
Stage-matched history
  & Failed & Failed & 0.732 \\
Clean history and future & 0.826 & 0.849 & 0.735   \\
Stage-matched history $+$ clean anchor KV (ours)
  & \textbf{0.838} & \textbf{0.859} & \textbf{0.753} \\
\bottomrule
\end{tabular*}
\end{center}
\end{table}

\paragraph{Conditioning modules.}
We compare the following cross-chunk conditioning designs:
(1) Stage-matched history;
(2) Clean history and future;
(3) Stage-matched history with clean history and future, our final design.
Variants are trained under the same setup and reported in Tab.~\ref{tab:conditioning-modules} (bottom).
With only stage-matched history, the generated videos flicker so heavily that VBench fails to score their temporal flickering dimension.
With only clean history and future, the model becomes a variant of Self-Forcing with a different temporal generation order.
It mitigates the drift issue of Self-Forcing but still generates at a high latency; hence, we only utilize this as a sparse conditioning module rather than a dense generation model.
By combining these modules, \method{} demonstrates better generation quality with faster generation speed.

\paragraph{Limitations and future work.}
\method{} targets long-sequence, few-step generation on multiple GPUs.
Planner and pipeline warm-up reduce its gains for short videos, while one-GPU inference cannot exploit the wavefront parallelism.
Our training data consists only of real-world videos, which biases the model toward photorealistic content and may limit its performance on animation and other stylized domains.
We use a fixed anchor stride, conditioning window, and vanilla training recipe except for the time-rebased augmentation; relaxing the current clean anchor-state assumption and adapting anchor selection remain future work.
The cross-chunk state contract may also complement KV selection and compression, RoPE~\citep{su2024roformer} modifications, and anti-drift techniques.
Parallel generation of anchor-delimited intervals~\citep{xiang2025mmpl} is also promising.

\section{Conclusion}

\method{} treats cross-chunk memory as a state-availability problem: a cross-chunk state contract specifies what KV is published and when a later chunk may consume it.
Publishing the KV produced by each ordinary renderer forward makes stage-matched renderer history available early enough to induce an inter-chunk wavefront over stage workers, so every renderer pass advances the output.
Because this promptly available history is noisy, fine-scale, and past-only, sparse auxiliary anchor latents are converted into clean anchor KV ahead of the local regions they guide, providing a complementary coarse structural reference.
The renderer uses this clean anchor KV only as conditioning and regenerates every final position.
This temporal-scale allocation enables inter-chunk stage pipelining.
Planner and renderer roles with a shared backbone realize the dependency graph through packed real-video supervision and full-clip self-rollout distillation.
Across 1.3B and 14B backbone scales at 480p and 720p, \method{} is the fastest evaluated autoregressive schedule on four GPUs for outputs of 321 RGB frames or more.
On four H100 GPUs with 1.3B model at $480$p, it generates videos faster than other methods with better quality and temporal coherence across several durations.

\bibliographystyle{assets/plainnat}
\bibliography{iclr2027_conference}

\clearpage
\newpage
\beginappendix

\paragraph{Appendix overview.}
The appendix is organized into three sections.
Sec.~\ref{app:latency-results} presents the batched multi-device wavefront algorithm; Sec.~\ref{app:anchor-rate} explains wavefront execution, anchor spacing, and planner--renderer rate matching, while Sec.~\ref{app:tables} documents the measurement protocol and planner allocations, analyzes latency and peak memory, analyzes the theoretical acceleration ratio with wall-clock break-even conditions, and evaluates streaming generation with a dedicated planner.
Sec.~\ref{app:regenerate-anchors} explains why the renderer regenerates anchor positions, using visual and quantitative evidence to separate the effects of the anchor-KV source and the final anchor source.
Sec.~\ref{app:training-details} gives the full training and implementation details: Sec.~\ref{app:supervised-details} covers packed time-rebased SFT and its optimization, whereas Sec.~\ref{app:distillation-details} describes self-rollout guidance, score tiling, gradient routing, and the optimization.

\emph{We provide an interactive demo on the \projectpage{}} that compares generation quality and latency with other methods, illustrates our generation process and training phases, and shows examples of the chunk-continuity and anchor--non-anchor seam issues that our framework resolves.

\section{Generation Efficiency Analysis}
\label{app:latency-results}

Throughout this appendix, a \emph{rank} is one GPU worker, and a renderer
\emph{round} is one synchronized pipeline interval in which every active rank
executes at most one chunk-sized forward at its assigned stage. The stage index
$k\in\{S,\ldots,1\}$ denotes the update from noise level $\sigma_k$ to
$\sigma_{k-1}$, while $\tau_k=\tau(\sigma_k)$ denotes the corresponding model
time input. We use \emph{context parallelism} (CP) for a different form
of parallelism: sharding one logical model forward across multiple ranks.

\subsection{Wavefront Execution and Anchor Spacing}
\label{app:anchor-rate}

\begin{algorithm}
\small
\caption{\textbf{Batched multi-device wavefront generation.}}
\label{alg:wavefront}
\begin{algorithmic}[1]
\Require Condition $c$; length $L$, positions $\mathcal{I}=\{0,\dots,L-1\}$; local
  anchor-window map $\Gamma$; roles
  $F^{\mathrm P}_{\theta},F^{\mathrm R}_{\theta}$.
\item[\textbf{Constants:}] $S{=}4$ stages with
  $\sigma_S>\cdots>\sigma_0{=}0$, $\Delta{=}10$, $B_{\mathrm P}{=}B_{\mathrm R}{=}3$,
  and a context window of $W_{\mathrm{ctx}}=21$ latent-frame positions for both roles;
  $H_{\mathrm P}{=}\lfloor(W_{\mathrm{ctx}}{-}B_{\mathrm P})/B_{\mathrm P}\rfloor{=}6$
  is the planner-history limit, and
  $H{=}\lfloor(W_{\mathrm{ctx}}{-}B_{\mathrm P}{-}B_{\mathrm R})/B_{\mathrm R}\rfloor{=}5$
  preceding renderer chunks fit in each stage-matched renderer history.
\item[\textbf{Layout:}] $M\gets\lfloor(L{-}1)/\Delta\rfloor+1$ anchors $\mathcal{P}$,
  $N_{\mathrm P}\gets\lceil M/B_{\mathrm P}\rceil$ blocks $\mathcal{Q}_j$, and
  $N_{\mathrm R}\gets\lceil L/B_{\mathrm R}\rceil$ chunks $\mathcal{C}_i$; the last block or chunk
  contains the remainder when necessary.
\item[\textbf{Device:}] Ranks $r\in\{0,\dots,S{-}1\}$; renderer rank $r$
  owns stage $k=S-r$. $\mathcal{B}^{(0)}_{k}\gets\emptyset,\mathcal{A}\gets\emptyset,\forall r$.
\item[\textbf{Phase A:}] Planning. \Comment{Complete the planner rollout before rendering, ending with $\mathcal{A}$ in host memory.}
\For{$j=0,\dots,N_{\mathrm P}-1$}
  \State Sample $x^{\mathrm P}_{j,S}\sim\mathcal{N}(0,1)$.
  \For{$k=S,\dots,1$}
    \State $\widehat v^{\mathrm P}_{j,k}\gets F^{\mathrm P}_{\theta}(x^{\mathrm P}_{j,k},c,\tau_k;
      \operatorname{Tail}_{H_{\mathrm P}}(\mathcal{A}))$; step
      $x^{\mathrm P}_{j,k-1}\gets x^{\mathrm P}_{j,k}-(\sigma_k{-}\sigma_{k-1})
      \widehat v^{\mathrm P}_{j,k}$.
  \EndFor
  \State $(K^{\mathrm P}_{j},V^{\mathrm P}_{j})\gets F^{\mathrm P}_{\theta}(x^{\mathrm P}_{j,0},c,\tau_0;\operatorname{Tail}_{H_{\mathrm P}}(\mathcal{A}))$;
    $\mathcal{A}\gets\operatorname{Append}(\mathcal{A},(K^{\mathrm P}_{j},V^{\mathrm P}_{j})),\forall j$.
    \Comment{including the last}
\EndFor
\item[\textbf{Phase B:}] Renderer wavefront, $N_{\mathrm R}+S-1$ rounds.
\For{$n=0,\dots,N_{\mathrm R}+S-2$}
  \ForAll{$r$ \textbf{in parallel}, with $i\gets n-r$ and $k\gets S-r$, active iff
    $0\le i<N_{\mathrm R}$}
    \State \emph{Obtain} $x^{\mathrm R}_{i,k}$: rank $0$ samples noise from $\mathcal{N}(0,1)$; rank $r{>}0$
      takes what rank $r{-}1$ sent in round $n{-}1$.
    \State \emph{Read} anchor KV $\mathcal{A}_{\Gamma(i)}$ and stage-matched history $\mathcal{B}^{(i)}_{k}$.
    \State \emph{Evaluate} Eq.~(\ref{eq:step-update}):
      $(\widehat v^{\mathrm R}_{i,k},K^{\mathrm R}_{i,k},V^{\mathrm R}_{i,k})\gets
      F^{\mathrm R}_{\theta}(x^{\mathrm R}_{i,k},c,\tau_k;\mathcal{A}_{\Gamma(i)},
      \mathcal{B}^{(i)}_{k})$.
    \State \emph{Step} $x^{\mathrm R}_{i,k-1}\gets x^{\mathrm R}_{i,k}
      -(\sigma_k{-}\sigma_{k-1})\widehat v^{\mathrm R}_{i,k}$.
    \State \emph{Write} in place:
    $\mathcal{B}^{(i+1)}_{k}\gets
      \operatorname{Tail}_{H}(\operatorname{Append}(\mathcal{B}^{(i)}_{k},
      (i,K^{\mathrm R}_{i,k},V^{\mathrm R}_{i,k})))$. \Comment{keep the most recent $H$ chunks}
    \State If $r<S{-}1$, \emph{send} $x^{\mathrm R}_{i,k-1}$ to rank $r{+}1$; otherwise
      \emph{keep} the finished $x^{\mathrm R}_{i,0}$ on $\mathcal{C}_i$.
  \EndFor
\EndFor
\Ensure The $L$ latents of $\mathcal{I}$, all produced by the renderer, on the last rank.
\end{algorithmic}
\end{algorithm}
\vspace{-0.5cm}

\paragraph{Wavefront generation.}
Alg.~\ref{alg:wavefront} describes the batched $S$-device setup used in our main latency matrix.
During generation, the devices first run the planner.
We could use only one rank to run the planner or use all ranks to run the planner with context parallelism.
We discuss the design choice later in details.
Either way, the planner will produce the clean anchor KV bank $\mathcal{A}$.
Each device then runs one renderer stage and stores its own cache bank.
After warm-up, the renderer chunk--stage wavefront in Eq.~(\ref{eq:step-update}) generates one chunk per round.

\paragraph{Anchor spacing and streaming rate.}
We keep the planner block size equal to the renderer chunk size, $B_{\mathrm P}=B_{\mathrm R}=3$, so that their forwards have comparable cost.
The stride $\Delta=10$ is chosen by matching anchor production and consumption in the streaming allocation with one planner device and four renderer devices.
One anchor block requires four denoising forwards and one clean-endpoint KV forward, or five planner rounds.
Each local three-anchor window then conditions six to seven renderer chunks, which the steady-state wavefront retires in six to seven rounds.
The planner can therefore publish the next anchors before the renderer exhausts its current window.
A smaller stride can make planning the bottleneck; a larger stride provides no latency gain once planning stays ahead, and leaves longer regions less constrained temporally.

\subsection{Latency and Peak Memory Analysis}
\label{app:tables}

\paragraph{Measurement protocol.}
Throughout the paper, reported latency and FPS refer to \emph{denoising-path}
latency.
All results in Tab.~\ref{tab:generation-time}, Tab.~\ref{tab:peak-memory}, and Fig.~\ref{fig:latency} use identical timing boundaries.
Timing spans the first planner or baseline denoising forward through the availability of completed video latents on the reporting rank.
Text encoding and VAE decoding are therefore excluded.
Planner denoising and clean-endpoint forwards, renderer forwards, cache operations,
pipeline synchronization and communication, and the role-weight transfer described
below are included whenever the corresponding schedule performs them.
Each cell reports the average over 20 generated videos after excluding the cold-start steps.
We reset the memory counters and synchronize all ranks before and after each timed window.
For each video, we report the slowest-rank elapsed time and the highest peak allocation across all ranks.
Among cells with per-video memory measurements, the standard deviation of peak memory has a median of 0.0000~GiB and a maximum of 0.4028~GiB, so we omit it from the memory table.
The three autoregressive methods, Self-Forcing, HiAR, and \method{}, use
three-latent chunks and a context budget of $W_{\mathrm{ctx}}=21$ latent-frame positions,
while the bidirectional baseline does not use caching.
For \method{}, these 21 positions comprise three clean anchor-KV positions in $\mathcal{A}_{\Gamma(i)}$, up to 15 renderer-history positions from $\mathcal{B}^{(i)}_{k}$, and up to $B_{\mathrm R}=3$ positions in the current chunk.
The planner uses the same budget: up to 18 completed anchor latents and up to $B_{\mathrm P}=3$ latents in its current block.
We evaluate the 1.3B and 14B models on H100 and GB300 GPUs, respectively.
The 14B rows are systems-only timing measurements: they instantiate the same
tensor shapes, masks, cache paths, and execution schedules with the 14B backbone.

\paragraph{Serving the two roles inside the timed window.}
Routing latents through adapters would add computation to every forward pass.
Therefore, for latency evaluation, we merge each
role's adapter into the backbone, producing two full-weight models.
A \method{} forward is then a standard backbone forward and has the same cost as a baseline forward on the same model.
This keeps the cost of a block-sized forward comparable across all three schedules.
To limit peak device memory, only the active model is kept on the device.
At the anchor-to-renderer boundary, the planner weights are off-loaded and the renderer weights are loaded from pinned host memory.
This transfer takes 0.10--0.13~s for the 1.3B model and 0.13--0.17~s for the 14B model.
The transfer time is included in the four-device \method{} latencies reported in Tab.~\ref{tab:generation-time} and Fig.~\ref{fig:latency}.
Streaming keeps both models resident on their assigned devices.

\paragraph{Planner allocations.}
We support two four-GPU planner allocations.
The default allocation (shown as \method{} in the tables) runs the planner on a single device.
\method{}-CP instead shards each logical planner forward across four devices
using context parallelism, which introduces collective-communication overhead.
In our experiments, the vanilla version is generally faster for 480p videos, whereas the context-parallel version is faster for 720p videos.
We report both variants so that users can choose the best deployment strategy for their target resolution.
Sec.~\ref{app:streaming} evaluates a five-GPU variant with a planner concurrent with the four-device renderer wavefront.

\begin{table}
\caption{Per-video denoising-path latency in seconds (lower is better).
Each cell is the mean over videos with its standard deviation beneath it.
Each setting spans four output lengths: 5, 20, 35, and 65 seconds, which are 21, 81, 141, and 261 latents at 16 FPS.
The two \method{} planner allocations are both measured in every four-GPU cell.
All multi-GPU columns use four GPUs unless marked otherwise; \method{}-Streaming uses one planner GPU and four renderer GPUs.
Boldface identifies the fastest schedule at an equal device count and therefore excludes the five-GPU row.
}
\label{tab:generation-time}
\begin{center}
\footnotesize
\setlength{\tabcolsep}{1.9pt}
\begin{tabularx}{\linewidth}{@{}l*{12}{Y}@{}}
\toprule
& \multicolumn{12}{c}{Wan2.1-T2V-1.3B (H100)} \\
\cmidrule(lr){2-13}
& \multicolumn{4}{c}{480p, $1{\times}$GPU} & \multicolumn{4}{c}{480p, multi-GPU}
& \multicolumn{4}{c}{720p, multi-GPU} \\
\cmidrule(lr){2-5}\cmidrule(lr){6-9}\cmidrule(lr){10-13}
Schedule
 & 5 & 20 & 35 & 65 & 5 & 20 & 35 & 65 & 5 & 20 & 35 & 65 \\
\midrule
\mslab{Bidirectional}
 & \ms{\textbf{3.31}}{0.00}& \ms{32.02}{0.01}& \ms{91.22}{0.04}& \ms{295.83}{0.12}
 & \ms{\textbf{1.13}}{0.01}& \ms{8.95}{0.01}& \ms{24.01}{0.02}& \ms{75.28}{0.05}
 & \ms{\textbf{3.74}}{0.00}& \ms{39.59}{0.06}& \ms{115.79}{0.04}& \ms{379.17}{0.64} \\
\mslab{Self-Forcing}
 & \ms{3.69}{0.01}& \ms{\textbf{17.44}}{0.11}& \ms{\textbf{31.48}}{0.14}& \ms{\textbf{57.89}}{0.20}
 & \ms{5.10}{0.05}& \ms{20.25}{0.10}& \ms{35.40}{0.31}& \ms{65.14}{0.55}
 & \ms{6.04}{0.05}& \ms{27.17}{0.11}& \ms{47.71}{0.27}& \ms{97.68}{0.48} \\
\mslab{HiAR}
 & \ms{5.57}{0.04}& \ms{27.86}{0.03}& \ms{50.41}{0.20}& \ms{94.74}{0.15}
 & \ms{2.06}{0.05}& \ms{7.67}{0.00}& \ms{13.31}{0.01}& \ms{24.67}{0.03}
 & \ms{6.49}{0.01}& \ms{26.82}{0.02}& \ms{46.94}{0.03}& \ms{86.85}{0.06} \\
\midrule
\mslab{\method{}}
 & \ms{4.56}{0.07}& \ms{17.77}{0.14}& \ms{32.80}{0.21}& \ms{59.64}{0.46}
 & \ms{2.21}{0.08}& \ms{\textbf{6.32}}{0.02}& \ms{\textbf{10.79}}{0.22}& \ms{\textbf{21.29}}{0.08}
 & \ms{5.71}{0.05}& \ms{19.33}{0.27}& \ms{35.14}{0.13}& \ms{66.53}{0.12} \\
\mslab{\method{}-CP}
 & - & - & - & -
 & \ms{2.34}{0.02}& \ms{7.24}{0.06}& \ms{12.30}{0.28}& \ms{22.07}{0.20}
 & \ms{5.50}{0.03}& \ms{\textbf{18.47}}{0.35}& \ms{\textbf{30.90}}{0.38}& \ms{\textbf{58.89}}{0.18} \\
\multicolumn{1}{@{}>{\columncolor{blue}[0pt][4pt]}l}{\shortstack[l]{\method{}-Streaming\\[-0.15em]{\tiny($5{\times}$GPU)}}}
 & \multicolumn{1}{>{\columncolor{blue}[4pt][4pt]}r}{-}
 & \multicolumn{1}{>{\columncolor{blue}[4pt][4pt]}r}{-}
 & \multicolumn{1}{>{\columncolor{blue}[4pt][4pt]}r}{-}
 & \multicolumn{1}{>{\columncolor{blue}[4pt][4pt]}r}{-}
 & \multicolumn{1}{>{\columncolor{blue}[4pt][4pt]}r}{\ms{1.85}{0.01}}
 & \multicolumn{1}{>{\columncolor{blue}[4pt][4pt]}r}{\ms{5.11}{0.01}}
 & \multicolumn{1}{>{\columncolor{blue}[4pt][4pt]}r}{\ms{8.27}{0.01}}
 & \multicolumn{1}{>{\columncolor{blue}[4pt][4pt]}r}{\ms{14.53}{0.02}}
 & \multicolumn{1}{>{\columncolor{blue}[4pt][4pt]}r}{\ms{5.47}{0.00}}
 & \multicolumn{1}{>{\columncolor{blue}[4pt][4pt]}r}{\ms{16.37}{0.02}}
 & \multicolumn{1}{>{\columncolor{blue}[4pt][4pt]}r}{\ms{27.05}{0.05}}
 & \multicolumn{1}{>{\columncolor{blue}[4pt][0pt]}r@{}}{\ms{48.61}{0.07}} \\
\bottomrule
\end{tabularx}
\begin{tabularx}{\linewidth}{@{}l*{8}{Y}@{}}
\toprule
& \multicolumn{8}{c}{Wan2.1-T2V-14B (GB300)} \\
\cmidrule(lr){2-9}
& \multicolumn{4}{c}{480p, $4{\times}$GPU} & \multicolumn{4}{c}{720p, $4{\times}$GPU} \\
\cmidrule(lr){2-5}\cmidrule(lr){6-9}
Schedule
 & 5 & 20 & 35 & 65 & 5 & 20 & 35 & 65 \\
\midrule
\mslab{Bidirectional}
 & \ms{\textbf{3.50}}{0.01}& \ms{31.83}{0.01}& \ms{90.23}{0.01}& \ms{295.45}{0.01}
 & \ms{\textbf{12.79}}{0.01}& \ms{154.47}{0.01}& \ms{454.53}{0.01}& \ms{1530.42}{0.03} \\
\mslab{Self-Forcing}
 & \ms{9.75}{0.09}& \ms{41.35}{0.15}& \ms{74.56}{0.17}& \ms{137.89}{0.37}
 & \ms{15.84}{0.03}& \ms{78.88}{0.07}& \ms{142.29}{0.11}& \ms{269.41}{0.20} \\
\mslab{HiAR}
 & \ms{5.64}{0.00}& \ms{22.79}{0.02}& \ms{39.63}{0.02}& \ms{73.03}{0.03}
 & \ms{21.72}{0.00}& \ms{93.25}{0.03}& \ms{163.66}{0.05}& \ms{304.35}{0.08} \\
\midrule
\mslab{\method{}}
 & \ms{4.79}{0.04}& \ms{\textbf{15.67}}{0.11}& \ms{\textbf{27.67}}{0.22}& \ms{53.43}{0.41}
 & \ms{17.19}{0.09}& \ms{61.39}{0.34}& \ms{109.89}{0.03}& \ms{214.83}{0.54} \\
\mslab{\method{}-CP}
 & \ms{5.23}{0.02}& \ms{16.66}{0.12}& \ms{28.10}{0.33}& \ms{\textbf{52.18}}{0.27}
 & \ms{16.28}{0.07}& \ms{\textbf{55.66}}{0.20}& \ms{\textbf{97.02}}{0.50}& \ms{\textbf{181.25}}{0.80} \\
\multicolumn{1}{@{}>{\columncolor{blue}[0pt][4pt]}l}{\shortstack[l]{\method{}-Streaming\\[-0.15em]{\tiny($5{\times}$GPU)}}}
 & \multicolumn{1}{>{\columncolor{blue}[4pt][4pt]}r}{\ms{4.72}{0.05}}
 & \multicolumn{1}{>{\columncolor{blue}[4pt][4pt]}r}{\ms{13.40}{0.02}}
 & \multicolumn{1}{>{\columncolor{blue}[4pt][4pt]}r}{\ms{22.35}{0.05}}
 & \multicolumn{1}{>{\columncolor{blue}[4pt][4pt]}r}{\ms{39.43}{0.32}}
 & \multicolumn{1}{>{\columncolor{blue}[4pt][4pt]}r}{\ms{17.44}{0.10}}
 & \multicolumn{1}{>{\columncolor{blue}[4pt][4pt]}r}{\ms{53.17}{0.14}}
 & \multicolumn{1}{>{\columncolor{blue}[4pt][4pt]}r}{\ms{88.79}{0.01}}
 & \multicolumn{1}{>{\columncolor{blue}[4pt][0pt]}r@{}}{\ms{160.34}{0.27}} \\
\bottomrule
\end{tabularx}
\end{center}
\end{table}
\begin{table}
\caption{Peak allocated memory per rank in GiB.
}
\label{tab:peak-memory}
\begin{center}
\footnotesize
\setlength{\tabcolsep}{3pt}
\begin{tabular*}{\linewidth}{@{\extracolsep{\fill}}lrrrrrrrrrrrr@{}}
\toprule
& \multicolumn{12}{c}{Wan2.1-T2V-1.3B (H100)} \\
\cmidrule(lr){2-13}
& \multicolumn{4}{c}{480p, $1{\times}$GPU} & \multicolumn{4}{c}{480p, $4{\times}$GPU}
& \multicolumn{4}{c}{720p, $4{\times}$GPU} \\
\cmidrule(lr){2-5}\cmidrule(lr){6-9}\cmidrule(lr){10-13}
Schedule
 & 5 & 20 & 35 & 65 & 5 & 20 & 35 & 65 & 5 & 20 & 35 & 65 \\
\midrule
Bidirectional
 & 22.1 & 31.7 & 41.2 & 60.3
 & 15.7 & 18.4 & 21.1 & 26.5
 & 17.0 & 23.2 & 29.4 & 41.8 \\
Self-Forcing
 & 24.9 & 24.9 & 25.0 & 25.1
 & 20.6 & 20.6 & 20.7 & 20.7
 & 28.1 & 28.2 & 28.3 & 28.5 \\
HiAR
 & 24.9 & 24.9 & 25.0 & 25.1
 & 24.9 & 24.9 & 25.0 & 25.1
 & 32.9 & 33.0 & 33.1 & 33.3 \\
\method{}
 & 24.9 & 25.0 & 25.2 & 25.4
 & 24.9 & 24.9 & 25.0 & 25.1
 & 32.9 & 33.0 & 33.1 & 33.3 \\
\bottomrule
\end{tabular*}

\begin{tabular*}{\linewidth}{@{\extracolsep{\fill}}lrrrrrrrr@{}}
\toprule
& \multicolumn{8}{c}{Wan2.1-T2V-14B (GB300)} \\
\cmidrule(lr){2-9}
& \multicolumn{4}{c}{480p, $4{\times}$GPU} & \multicolumn{4}{c}{720p, $4{\times}$GPU} \\
\cmidrule(lr){2-5}\cmidrule(lr){6-9}
Schedule
 & 5 & 20 & 35 & 65 & 5 & 20 & 35 & 65 \\
\midrule
Bidirectional
 & 28.8 & 34.0 & 39.1 & 49.4
 & 31.2 & 43.1 & 54.9 & 78.7 \\
Self-Forcing
 & 60.1 & 60.2 & 60.2 & 60.3
 & 93.2 & 93.3 & 93.4 & 93.6 \\
HiAR
 & 79.0 & 79.0 & 79.1 & 79.1
 & 112.8 & 112.9 & 113.0 & 113.2 \\
\method{}
 & 79.0 & 79.0 & 79.1 & 79.2
 & 112.8 & 112.9 & 113.0 & 113.3 \\
\bottomrule
\end{tabular*}
\end{center}
\end{table}

\subsubsection{Analysis}

\paragraph{Generation latency.}
Among the four-GPU schedules in Tab.~\ref{tab:generation-time}, the faster of the two \method{} planner allocations is the fastest in every model--resolution setting from 20 seconds onward, and it retains this lead at both 35 and 65 seconds.
At 65 seconds, \method{} reduces latency by $3.54$--$8.44\times$ relative to bidirectional generation, $1.49$--$3.06\times$ relative to Self-Forcing, and $1.16$--$1.68\times$ relative to HiAR across these settings.
The preferred planner allocation follows the workload: vanilla is generally faster at 480p, whereas context parallelism consistently wins at 720p.
Before amortization at 5 seconds, the bidirectional baseline remains fastest in all five settings.

On a single GPU, where the renderer cannot exploit the inter-chunk stage pipeline, \method{} is nevertheless $1.54$--$1.59\times$ faster than HiAR for 20--65-second outputs because it removes most of HiAR's cache-update-only forwards.
Compared with Self-Forcing, \method{} also reduces cache-update-only forwards, but it materializes and stores renderer-history KV at all four denoising stages, rather than once per chunk; these writes occur inside output-advancing forwards but still carry bookkeeping cost.
These two effects largely offset each other, leaving \method{} close to Self-Forcing in latency and only $1.9$--$4.2\%$ slower.

\paragraph{Peak memory.}
Tab.~\ref{tab:peak-memory} reports the memory allocation.
Bidirectional and Self-Forcing require less per-rank memory thanks to context parallelism.
Because the autoregressive methods use a fixed sliding window, their peak memory is nearly independent of video length.
The reported \method{} allocation remains within 0.3~GiB of HiAR's in every cell.

\subsubsection{Discussion: When Is the Pipeline Faster?}

Beyond the empirical latency results above, we provide a complementary theoretical comparison of the three autoregressive schedules.
Our goal is to characterize the conditions under which \method{} is expected to be faster.
We begin with their forward counts.

\begin{figure}[t]
\centering
\includegraphics[width=\linewidth]{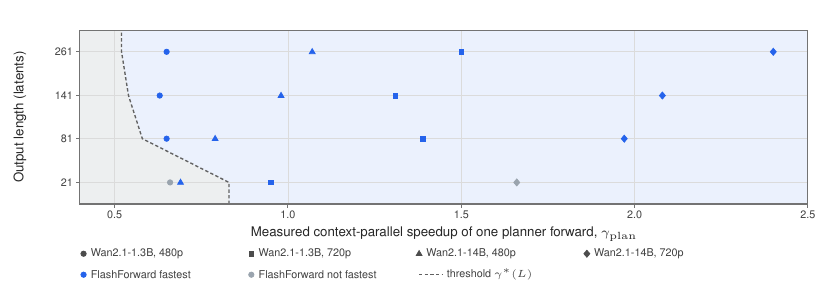}
\caption{Forward-count break-even boundary between \method{} and HiAR when the planner uses context parallelism, evaluated in the sixteen four-GPU cells of Tab.~\ref{tab:generation-time}.
Marker shape denotes the system setting; marker color reports whether \method{} is the fastest measured schedule among \method{}, HiAR, and Self-Forcing, using the same two colors as the predicted regions containing the markers.
The two gray markers are the cells in which \method{} does not win.
A marker whose color disagrees with its region is a cell in which the pairwise boundary alone does not recover the three-way winner.
Both outliers occur at a length of 21 latents, suggesting that \method{} is not well amortized for short sequences in these settings.
}
\label{fig:phase}
\end{figure}

\paragraph{Forward-count model.}
We compare the three autoregressive schedules in \emph{logical block-sized
forward equivalents}.  Let
$N_{\mathrm P}=\lceil M/B_{\mathrm P}\rceil$ be the number of planner blocks,
$N_{\mathrm R}=\lceil L/B_{\mathrm R}\rceil$ the number of renderer chunks, and
$N_{\mathrm C}=\lceil L/B_{\mathrm C}\rceil$ the number of chunks for either
autoregressive baseline.  All schedules use
$B_{\mathrm C}=B_{\mathrm R}=3$.  For $S$ denoising stages, their full-video
logical work is
\begin{equation}
\begin{aligned}
W_{\mathrm{\method{}}}&=N_{\mathrm P}(S+1)+N_{\mathrm R}S, \\
W_{\text{Self-Forcing}}&=N_{\mathrm C}S+(N_{\mathrm C}-1), \\
W_{\mathrm{HiAR}}&=N_{\mathrm C}S+(N_{\mathrm C}-1)S.
\end{aligned}
\label{eq:forward-count}
\end{equation}
The extra $1$ per planner block accounts for clean-KV construction, while the terminal cache-update-only forward is omitted for the baselines because it has no consumer.
At 81 latents, $N_{\mathrm P}=3$, $N_{\mathrm R}=N_{\mathrm C}=27$, and $S=4$, giving 123, 134, and 212 forwards, respectively.
On four devices, \method{} and HiAR assign different denoising stages to
different ranks, whereas Self-Forcing remains a serial chain whose individual
logical forwards use context parallelism; \method{}'s planner also remains
serial across anchor blocks.
We therefore use two workload-specific CP speedups.  Let
$\gamma_{\mathrm{plan}}$ be the latency of one complete one-rank planner
rollout divided by that of its four-rank CP rollout, and let
$\gamma_{\mathrm{SF}}$ be the latency of a one-rank Self-Forcing logical forward
divided by that of its four-rank CP counterpart under the same setting.

\paragraph{When does the forward reduction become a wall-clock speedup?}
 Including the $S-1$ wavefront
fill/drain slots, the loads and two break-even conditions on $S$-GPUs are
\begin{equation}
\begin{aligned}
R_{\mathrm{FF}}&=N_{\mathrm R}+\frac{N_{\mathrm P}(S+1)}{\gamma_{\mathrm{plan}}}+(S-1),\\
R_{\mathrm{SF}}&=\frac{N_{\mathrm C}S+(N_{\mathrm C}-1)}{\gamma_{\mathrm{SF}}},\\
R_{\mathrm{HiAR}}&=2N_{\mathrm C}-1+(S-1),\\
R_{\mathrm{FF}}<R_{\mathrm{HiAR}}
&\iff
\gamma_{\mathrm{plan}}>\frac{N_{\mathrm P}(S+1)}{2N_{\mathrm C}-1-N_{\mathrm R}},\\
R_{\mathrm{FF}}<R_{\mathrm{SF}}
&\iff
\gamma_{\mathrm{SF}}<\frac{N_{\mathrm C}S+(N_{\mathrm C}-1)}{R_{\mathrm{FF}}}.
\end{aligned}
\label{eq:phase-loads}
\end{equation}
Against HiAR, the required $\gamma_{\mathrm{plan}}$ falls from 0.83 at 21 latents to 0.58, 0.54, and 0.52 at 81, 141, and 261 latents.
As for $\gamma_{\mathrm{plan}}<1$ we could simply use one rank to run the planner and degenerates $\gamma_{\mathrm{plan}}=1$, we can always has less loads for model forwards compared with HiAR.
Against Self-Forcing, even the conservative choice $\gamma_{\mathrm{plan}}=1$ allows $\gamma_{\mathrm{SF}}$ to rise from 2.27 to 2.98, 3.12, and 3.21 over the same lengths before the ordering reverses.
Thus we can beat Self-Forcing in a large interval.

Fig.~\ref{fig:phase} compares this prediction with the measurements in Tab.~\ref{tab:generation-time}.
For each model, resolution, and output length, a point to the right of the dashed boundary predicts that \method{} is faster than HiAR, while a point to the left predicts the reverse; marker color reports whether \method{} is the fastest measured schedule of the three, so a marker whose color disagrees with the background region containing it is a cell in which the pairwise boundary alone does not recover the three-way winner.
The pairwise \method{}--HiAR prediction is correct in fifteen of the sixteen cells.
When all three schedules are compared, \method{} is fastest in fourteen cells: the only exceptions are the two 21-latent settings in which HiAR wins at 1.3B 480p and Self-Forcing wins at 14B 720p.
From 81 latents onward, \method{} is fastest in every model--resolution setting.

As videos become longer and each forward becomes more compute-intensive through a larger model or higher resolution, the measured context-parallel speedup tends to improve, while the break-even requirements above become less restrictive and fixed pipeline costs are better amortized.
\textbf{\emph{The conclusion that \method{} is faster than both baselines is therefore most robust for long, large-model, and high-resolution generation}}.
For small forwards with $\gamma_{\mathrm{plan}}<1$, the same analysis gives a direct deployment rule: use the planner on one rank rather than use context parallelism.

\paragraph{From forward counts to wall-clock speedup.}
The forward-count ratio is an upper-bound estimate rather than an exact latency prediction: in practice, context parallelism provides sublinear speedup because of collective overhead, and the wavefront pays fixed fill, drain, and synchronization costs.
The inter-stage latent tensor is small and its transfer overlaps the next forward, so communication volume is not the dominant term.
At 81 latents, the raw counts suggest a $212/123=1.72\times$ advantage over HiAR, while the measured speedup is $1.21$--$1.68\times$ across the four settings.
The gap is largest for short, low-compute workloads, but these costs are progressively amortized as the workload grows; at 14B and 720p, the measured $1.68\times$ speedup approaches the theoretical $1.72\times$, showing that the forward-count advantage translates into practical gains.

\subsubsection{Streaming Generation with a Dedicated Planner}
\label{app:streaming}

The four-device configuration in our main latency matrix reuses the same devices for both roles: all four first complete the sparse-anchor rollout and are then assigned to the four stages of the renderer wavefront to make a fair comparison with other autoregressive methods.
This serial planner prologue increases the time to the first completed latent chunk (our TTFF boundary), but it is an implementation choice rather than a dependency imposed by \method{}: a renderer chunk needs only its local anchor window, not anchors from later windows.
For deployment, we therefore dedicate one GPU to the planner and keep four GPUs on the four-stage renderer wavefront.
As soon as the planner publishes the first anchor block, rendering can begin; the planner then produces later blocks concurrently with rendering.
This reduces the anchor contribution to TTFF from a complete planner rollout to one block and, more importantly, supports streaming generation.

\paragraph{Measured five-GPU latency.}
We measure this one-planner--four-renderer schedule on both models, as reported in the \method{}-Streaming ($5{\times}$GPU) rows of Tab.~\ref{tab:generation-time}.
Here we compare against the four-GPU \method{} row, which
uses the same planner algorithm without CP; the comparison adds one device and
measures the deployment benefit of overlapping that planner with rendering.
At 21 latents (5 seconds), there is only one anchor block and hence no later planner work to overlap.  
The differences at this length therefore reflect model residency and, for the five-GPU 14B setup, cross-node communication rather than pipeline overlap.
As sequence length increases, the overlap benefit dominates.  
From 20 to 65 seconds, the dedicated H100 planner makes 1.3B generation $1.18$--$1.47\times$ faster at both 480p and 720p.
The 14B configuration spans two GB300 nodes, so each anchor block crosses the inter-node link and communication time grows with sequence length, but most anchor generation remains hidden behind rendering: relative to the four-GPU single-planner row, 14B generation is $1.15$--$1.36\times$ faster over the same lengths.
In both deployments, the overlap leaves renderer peak memory essentially unchanged from the standard four-GPU version.
Together, these results show that a dedicated producer converts the planner prologue into a one-block streaming warm-up and yields increasing denoising-path latency gains as the pipeline is amortized.

\paragraph{Only the first anchor block is exposed.}
To test whether the planner lies on the steady-state critical path, we separately record the producer's full anchor time and \texttt{anchor\_wait}, the time for which renderer ranks wait for unavailable anchors.
At 480p, the full anchor phase grows from 0.510 to 5.617~s as the number of planner blocks grows from one to nine, while \texttt{anchor\_wait} remains between 0.506 and 0.509~s.
At 720p, the full phase grows from 1.067 to 17.956~s, while the wait remains between 1.051 and 1.066~s.
Thus the exposed fraction of planning falls from 99--100\% for a single block to 9\% at 480p and 6\% at 720p for nine blocks.
The wait is always approximately the cost of the first block: after startup, the planner stays ahead and all later anchor generation is hidden beneath the renderer wavefront, as predicted by the rate-matching analysis in Sec.~\ref{app:anchor-rate}.

\section{Why the Renderer Regenerates the Anchor Positions}
\label{app:regenerate-anchors}

\paragraph{Cheaper variants that do not work.}
A cheaper design would keep the planner anchors in the final video and render only the positions between them.
We test two versions of this splice design.
The first reuses the KV produced by the planner, while the second passes the clean auxiliary anchor latents through the renderer to build renderer-side KV.
Both versions mix two output sources: planner latents at the anchor positions and renderer latents everywhere else.
Our method instead uses the planner anchors only as conditioning and lets the renderer generate every output position.
This comparison separates the effect of the anchor KV source from the effect of the final output source.

\paragraph{Experiments.}
To evaluate these choices, we run the experiments summarized in Tab.~\ref{tab:anchor-mismatch}.
All experiments use the same training setup as our final model.
Panel (a) compares the splice variants with all-position rendering.
Panel (b) measures how the mean luminance levels of planner anchor frames and neighboring renderer frames change across training checkpoints for these splice variants.
Panel (c) replaces generated planner anchors with latents encoded from real videos to test whether planner generation error alone explains the seam.

\begin{figure}
\centering
\includegraphics[width=\linewidth]{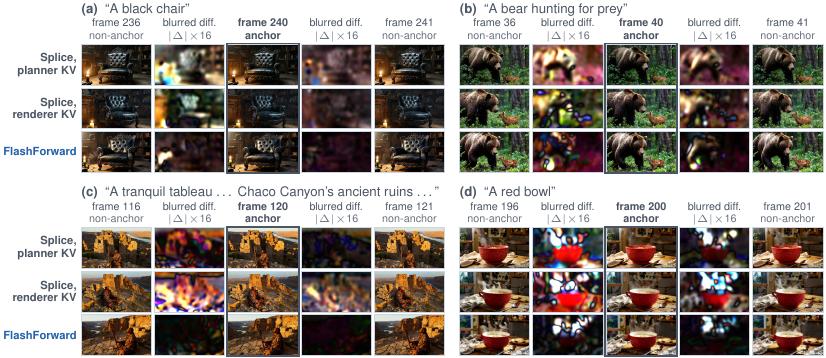}
\caption{
\textbf{Color seams at the anchor positions of 20-second videos.}
The two splice rows keep the planner anchor latents in the final video and build the anchor KV with the planner or with the renderer, as in Panel (a) of Tab.~\ref{tab:anchor-mismatch}; \method{} uses the planner anchors only as conditioning and lets the renderer generate every output position.
In each block, the boxed column is the last frame decoded from an anchor latent, and the outer columns are the nearest non-anchor frames before and after it.
Between each non-anchor frame and the anchor frame, we show their per-channel absolute RGB difference, computed after a Gaussian blur and amplified $16\times$.
The blur suppresses edges and fine texture, leaving mostly low-frequency color changes; large moving regions can remain visible.
}
\label{fig:anchor-seam}
\end{figure}

\paragraph{What the seam looks like.}
Fig.~\ref{fig:anchor-seam} visualizes this comparison on four 20-second videos, using the same frame indices for all three variants.
In (a) and (b), the anchor frames of both splice variants are darker than their neighbors; under renderer-side KV, the rocks in (c) turn lighter and warmer; and in (d), the bowl turns a brighter red.
The blurred differences make the offset visible.
For both splice variants, the differences highlight the static background as well as the moving subject, indicating a color change across the whole frame rather than motion.
For \method{}, they stay dark except where large regions move, such as the candle smoke in (a), the bear's body in (b), and the steam above the bowl in (d).
The KV source changes how strongly the offset appears in each clip, but neither splice variant removes it, consistent with Panel (a).
We \emph{include videos of the two variants on the \projectpage[\#issues]{}}; they show more directly what the seam looks like in motion.

\begin{table}
\caption{
\textbf{All-position renderer output leads to smaller seam measurements.}
\emph{Panel (a)} compares anchor-KV and final-anchor sources.
All seam columns are normalized adjacent-frame luminance-jump statistics, and lower is better: Anchor seam is measured at anchor/non-anchor transitions, Chunk seam at renderer-chunk boundaries away from anchors, and Interior at transitions of neither type.
``Periodic'' sums the excess over Interior across the transitions in one anchor period.
\emph{Panel (b)} reports mean luminance levels for planner-generated anchor frames and their neighboring renderer-generated frames.
\emph{Panel (c)} compares real and generated anchors over paired clips; chromatic share is the fraction of the RGB offset orthogonal to the grayscale direction $(1,1,1)$, and spatial excess is $\sqrt{s_{\mathrm{anchor}}^{2}-s_{\mathrm{control}}^{2}}$, where $s$ is the spatial standard deviation of the per-cell RGB change.
The paired $t$-statistic compares each clip's anchor-boundary offset with its mid-chunk control.
}
\label{tab:anchor-mismatch}
\center{
\scriptsize
\setlength{\tabcolsep}{3pt}
\begin{tabular*}{\linewidth}{@{\extracolsep{\fill}}rllrrrr@{}}
\multicolumn{7}{@{}l}{\textbf{(a) Compared modes: KV source and final anchor source}} \\[0.2em]
\toprule
Denoising steps & Final anchor & Anchor KV & Anchor seam & Chunk seam & Interior & Periodic \\
\midrule
50 & Planner & Planner  & 3.93 & 2.71 & 1.32 & 7.66 \\
50 & Planner & Renderer & 3.34 & 2.37 & 1.29 & 5.99 \\
50 & Renderer (ours)  & Planner  & \textbf{1.83} & 2.27 & 1.32 & \textbf{3.54} \\
4  & Planner & Renderer & 6.20 & 1.68 & 1.40 & 9.55 \\
4  & Renderer (ours)  & Planner  & \textbf{2.43} & 2.50 & 1.26 & \textbf{5.61} \\
\bottomrule
\end{tabular*}

\vspace{0.6em}
\begin{tabular*}{\linewidth}{@{\extracolsep{\fill}}lrrr@{}}
\multicolumn{4}{@{}l}{\textbf{(b) Seam formation when planner anchors are used as final outputs}} \\[0.2em]
\toprule
Checkpoint & Anchor level & Renderer level & Anchor $-$ renderer \\
\midrule
50-step supervised                    &  87.804 &  87.898 & $-0.094$ \\
4-step distilled, backbone frozen     &  98.192 &  99.364 & $-1.172$ \\
4-step distilled, backbone updated    & 130.043 & 136.287 & $\mathbf{-6.244}$ \\
\quad Paired change: updated $-$ frozen & $+31.85$ & $+36.92$ & $-5.07$ \\
\bottomrule
\end{tabular*}

\vspace{0.6em}
\begin{tabular*}{\linewidth}{@{\extracolsep{\fill}}lrrrrr@{}}
\multicolumn{6}{@{}l}{\textbf{(c) Compared modes: real anchors and generated planner anchors in spliced output}} \\[0.2em]
\toprule
Anchor source & Anchor offset & Mid-chunk control & Chromatic share & Spatial excess & Paired $t$-statistic  \\
\midrule
Real-video anchors             & 9.47 & 0.52 & 49\% & 6.69 & 4.34   \\
Generated planner anchors   & 3.00 & 0.37 & 47\% & 4.62 & 11.51  \\
\bottomrule
\end{tabular*}
}
\end{table}

\paragraph{At 50 steps, the final output source has the larger measured effect.}
Panel (a) gives the direct comparison.
At 50 steps, replacing planner KV with renderer-side KV only modestly reduces the anchor seam and periodic total, and the seam remains visible.
These one-factor comparisons show that changing the final anchor source has the larger measured effect at this phase.
In the deployed four-step comparison, all-position rendering again cuts the anchor seam by more than half relative to the splice configuration and clearly lowers the periodic total.

\paragraph{Planner--renderer output levels diverge after distillation.}
A splice baseline works only if planner anchors and renderer frames have matching output statistics.
Panel (b) shows that they nearly match after supervised fine-tuning, with a negligible mean luminance gap.
Four-step distillation widens this gap by more than an order of magnitude even with a frozen backbone, and updating the shared backbone widens it several times further.
Relative to the frozen model, updating the backbone raises both levels, but the renderer level rises more than the anchor level, so the anchor path lags behind the renderer's shift.
This unequal shift becomes visible when planner anchors are inserted into output.

\paragraph{The mismatch is not simply poor planner output.}
Panel (c) shows that the seam remains when generated planner anchors are replaced with anchors encoded from real videos.
The real-anchor probe removes planner generation error, yet all clips still have more error at the anchor boundary than at the mid-chunk control.
The mismatch also has more than one component: 47--49\% of the frame-level offset is chromatic, and the spatially varying excess remains large for both real and generated anchors.
These results are consistent with an output-interface mismatch rather than an artifact explained solely by a bad planner sample or a single brightness offset.

\paragraph{Why all-position rendering avoids cross-role splicing.}
Our method does not require planner and renderer outputs to match at a splice boundary.
The planner anchors provide clean structural guidance, but they are not copied into the final video.
The renderer generates every output latent, including the anchor positions, so the final sequence never switches between planner and renderer outputs.
This removes the cross-role output switch as one possible source of a periodic seam while preserving the anchors as conditioning, and thus leads to a substantial reduction of anchor seams.

\section{Training and Implementation Details}
\label{app:training-details}

Given a clean latent $x_0$, Gaussian noise $\varepsilon$, and a noise level $\sigma$, Rectified Flow~\citep{liu2023rectifiedflow} uses
\begin{equation}
x_{\sigma}=(1-\sigma)x_0+\sigma\varepsilon,
\qquad
v=\varepsilon-x_0.
\label{eq:flow-parameterization}
\end{equation}
We use a flow shift of 5.0 at every denoising stage.
We reserve $k\in\{S,\ldots,1\}$ for a discrete denoising-stage index.
For any noise level $\sigma$, let $\tau(\sigma)$ denote the scalar model time
input produced by the Wan scheduler with flow shift 5.0; at an inference stage,
$\tau_k=\tau(\sigma_k)$.
During distillation and inference, the student generator follows the four-stage guidance-free trajectory.
Only the frozen real-data score uses classifier-free guidance~\citep{ho2022cfg}, with a scale of 3.0, to construct the distillation target.
\emph{We provide an interactive demo on the \projectpage[\#training]{} to illustrate inputs and training phases.}

\subsection{Phase 1: Time-Rebased SFT and Optimization}
\label{app:supervised-details}

\begin{algorithm}[t]
\small
\caption{\textbf{Packed SFT forward.}}
\label{alg:sft}
\begin{algorithmic}[1]
\Require Captioned real videos; frozen VAE $\mathcal{E}$ and text encoder; trainable parameterization
  $\theta$ comprising the shared backbone, rank-$256$ role adapters $\psi_{\mathrm P},\psi_{\mathrm R}$, and two zero-initialized
  role-embedding vectors.
\item[\textbf{Constants:}] $L{=}81$, $\mathcal{P}{=}\{0,10,\dots,80\}$, $H{=}5$,
  $\Gamma(i)$,
  flow shift $5.0$.
\item[\textbf{Mask:}] Precompute the attention mask $\mathbf M$ with the following visible keys:
  \Statex \hskip\algorithmicindent
  $\mathrm{Vis}(\mathcal{P}^{\rho}[\mathcal{Q}_j])=
   \mathcal{P}^{\rho}[\mathcal{Q}_j]\cup
   \mathcal{P}^{0}[\bigcup_{q<j}\mathcal{Q}_q]$, $\rho\in\{0,\sigma\}$; \hskip\algorithmicindent \Comment{same for rebased anchors}
  \Statex \hskip\algorithmicindent
  $\mathrm{Vis}(\mathcal{I}^{\sigma}[\mathcal{C}_i])=
   \mathcal{I}^{\sigma}[\bigcup_{q=\max(0,i-H)}^{i}\mathcal{C}_q]\cup
   \mathcal{P}^{0}[\Gamma(i)]$.
\For{each training step}
  \State \emph{Encode} $x_0\gets\mathcal{E}(\text{clip})$ on $\mathcal{I}$; $c\gets$ caption, with
    $10\%$ text dropout.
  \State \emph{Rebase:} Sample distinct $o_1,\dots,o_4\in\{1,\dots,9\}$; for each $n$,
    reset $\mathcal{E}$, $\tilde x^{(n)}_0\gets\mathcal{E}(\text{clip}[4o_n,4o_n{+}281))$,
    keep its $8$ anchors at indices $\{0,10,\dots,70\}$.
  \State \emph{Pack} $z$ from planner copies $\mathcal{P}^{0},\mathcal{P}^{\sigma}$, the renderer copy
    $\mathcal{I}^{\sigma}$, and rebased planner copies
    $\{(\widetilde{\mathcal{P}}^{0}_{n},\widetilde{\mathcal{P}}^{\sigma}_{n})\}_{n=1}^{4}$.
    \Comment{a clean copy only where a later block reads it}
  \State \emph{Noise:} Sample one $\sigma$ per clip and set its model time input
    $\tau_{\sigma}\gets\tau(\sigma)$;
    $x_\sigma\gets(1{-}\sigma)x_0+\sigma\varepsilon$; clean
    copies stay at $\sigma_0{=}0$. \Comment{copies of one latent share $\varepsilon$}
  \State \emph{Route} planner copies via $\psi_{\mathrm P}$ and renderer copies via $\psi_{\mathrm R}$;
    add the corresponding role embedding to AdaLN.
    \Comment{one $\theta$ gives $F^{\mathrm P}_{\theta}$ and $F^{\mathrm R}_{\theta}$}
  \State \emph{Forward} $(\widehat v^{\mathrm P},\widehat v^{\mathrm R},
    \{\widehat{\tilde v}^{\mathrm P}_{n}\})\gets
    F_{\theta}(z,c,\tau_{\sigma};\mathbf M)$ on noised tokens; clean copies provide KV only.
  \State \emph{Score} $v\gets\varepsilon-x_0$ (Eq.~(\ref{eq:flow-parameterization}));
    $\ell\gets\lVert\widehat v-v\rVert^{2}$; average $\bar{\ell}_{\mathrm P}$,
    $\bar{\ell}_{\mathrm R}$, $\bar{\ell}^{\,\mathrm{reb}}_{\mathrm P}$ over the corresponding inputs;
  \Statex \hskip\algorithmicindent
  $\displaystyle
   \mathcal{L}\gets
   \bar{\ell}_{\mathrm P}+\bar{\ell}_{\mathrm R}
         +\bar{\ell}^{\,\mathrm{reb}}_{\mathrm P}$.
  \State \emph{Update:} Backpropagate once; take an AdamW step; update the EMA at decay $0.999$.
\EndFor
\Ensure The trained model and its EMA copy.
\end{algorithmic}
\end{algorithm}

\paragraph{Packed input sequence and attention mask.}
Alg.~\ref{alg:sft} packs both roles into one forward.
For a clip of $L=81$ latents, the planner uses the $M=9$ anchors $\mathcal{P}=\{0,10,\ldots,80\}$ in three blocks of three latents, while the renderer treats all $81$ positions as targets in $27$ chunks of three and reads up to $5$ preceding chunks.
Each anchor appears as a clean planner-context copy in $\mathcal{P}^{0}$, a noised planner input in $\mathcal{P}^{\sigma}$, and a noised renderer input in $\mathcal{I}^{\sigma}$; every non-anchor appears only as a renderer input in $\mathcal{I}^{\sigma}$.
All noised copies use the clip's single sampled $\sigma$, and copies of the same latent share $\varepsilon$.
The attention mask isolates the planner, renderer, and time-rebased sequences:
planner inputs attend within their current block and to clean earlier planner blocks, whereas each renderer chunk attends within the chunk, to up to five preceding renderer chunks at the same noise level, and to the clean anchors selected by $\Gamma(i)$.
The renderer's read of these clean anchors is the only cross-role connection.

\paragraph{Training on 20-second real videos.}
Real-video supervision is essential for learning the planner's autoregressive dependence across anchor blocks.
Under a 5-second training setup, each clip contains only three anchor latents, thus provides no supervision for autoregressive generation.
Distillation-based methods learn from trajectories produced by a teacher model; unless their training pipelines are substantially redesigned, their supervision horizon is typically limited to the teacher's 5-second generation window~\citep{yin2025causvid,huang2025selfforcing}.
Because our SFT phase learns directly from real videos rather than teacher trajectories, it can use clips of arbitrary duration.
Balancing stronger long-horizon supervision against training efficiency, we use 20-second clips, which contain nine anchors and form exactly three anchor blocks, ensuring that the planner's autoregressive generation is explicitly trained on real data.

\paragraph{Time-rebased planner supervision.}
At each step, we sample four distinct latent-time offsets $o_1,\ldots,o_4$ from $\{1,\ldots,9\}$.
Since the causal VAE downsamples time by four, offset $o_n$ defines the $281$-frame crop $[4o_n,4o_n+281)$.
The resulting $71$ latents provide eight planner targets at local positions $\{0,10,\ldots,70\}$.
These targets form blocks of sizes $3+3+2$.
Only the first two blocks have clean context copies because the final block has no later consumer, so each rebased crop adds eight noised inputs and six clean copies and remains isolated from the other packed sequences.

\paragraph{Objective and optimization.}
The shared backbone is initialized from Wan2.1-T2V-1.3B~\citep{wan2025}; its planner and renderer rank-$256$ LoRA~\citep{hu2022lora} adapters act on the self-attention query, key, value, and output projections and on the cross-attention query and output projections. One zero-initialized vector per role is selected from a role-embedding table and added through adaptive layer normalization (AdaLN). The VAE encoder remains frozen. We apply per-token squared error to the rectified-flow velocity $v=\varepsilon-x_0$ and sum the averages over the $9$ original planner latents, $81$ renderer latents, and $32$ time-rebased planner latents, respectively. Supervised training uses AdamW~\citep{loshchilov2019adamw} for $1{,}800$ steps of global batch size $128$, $\beta_1=0$, $\beta_2=0.999$, weight decay $0.01$, and gradient clipping at $1.0$. The learning rate warms up to $10^{-5}$ over the first $100$ steps and then remains constant, and the exponential-moving-average (EMA) decay is $0.999$.

\subsection{Phase 2: Self-Rollout Guidance and Step Distillation}
\label{app:distillation-details}

\paragraph{Self-rollout and gradient routing.}
Phase 2 starts from the Phase-1 student and retains the same planner--renderer
layout.  Its four guidance-free denoising stages select the following base
scheduler indices, ordered from highest to lowest noise:
\begin{equation}
\mathcal{T}_{\mathrm{student}}=(q_4,q_3,q_2,q_1)=(999,937,833,624).
\label{eq:student-schedule}
\end{equation}
\begin{algorithm}[t]
\small
\caption{\textbf{Self-rollout distribution-matching distillation.}}
\label{alg:dmd}
\begin{algorithmic}[1]
\Require Trainable SFT student; frozen real-data score $F^{\mathrm{real}}$ (Wan2.1-T2V-14B);
  critic $F^{\mathrm{fake}}_{\phi}$ (Wan2.1-T2V-1.3B, with a trainable rank-$256$ LoRA); frozen VAE $\mathcal{E}$,
  $\mathcal{D}$.
\item[\textbf{Constants:}] $L{=}81$, $M{=}9$ anchors, $N_{\mathrm P}{=}3$,
  $B_{\mathrm P}{=}3$, $N_{\mathrm R}{=}27$, $B_{\mathrm R}{=}3$, $S{=}4$,
  $H{=}5$, $W_{\mathrm{score}}{=}21$.
\Function{Rollout}{$c$}
  \State $\mathcal{A}\gets\emptyset$.
  \State \emph{Exit:} Sample $k^\star\sim\mathcal{U}\{1,\dots,S\}$;
    stages above $k^\star$ run detached.
    \Comment{the rollout stops at $k^\star$}
  \For{$\mathcal{Q}_j=\{0,10,20\},\{30,40,50\},\{60,70,80\}$}
    \State $x^{\mathrm P}_{j,S}\gets$ noise on $\mathcal{Q}_j$.
    \For{$k=S,\dots,k^\star$}
      \State $\widehat v^{\mathrm P}_{j,k}\gets F^{\mathrm P}_{\theta}(x^{\mathrm P}_{j,k},c,\tau_k;
        \sg{\mathcal{A}})$.
      \State If $k>k^\star$, step to $x^{\mathrm P}_{j,k-1}$.
    \EndFor
    \State $\hat x^{\mathrm P}_{j}\gets x^{\mathrm P}_{j,k^\star}
      -\sigma_{k^\star}\widehat v^{\mathrm P}_{j,k^\star}$; run a cache-extraction forward at $\sigma_0{=}0$;
      $\mathcal{A}\gets\operatorname{Append}(\mathcal{A},(K^{\mathrm P}_{j},V^{\mathrm P}_{j}))$.
  \EndFor
  \State \emph{Pack} the $L$ samples from $\mathcal{N}(0,1)$, under the
    block-causal mask of Alg.~\ref{alg:sft}, with $\hat x^{\mathrm P}$
    entering through $\mathcal{A}$.
  \For{$k=S,\dots,k^\star$}
    \State In one packed forward at $\tau_k$, evaluate
      $\widehat v^{\mathrm R}_{i,k}\gets
      F^{\mathrm R}_{\theta}(x^{\mathrm R}_{i,k},c,\tau_k;
      \mathcal{A}_{\Gamma(i)},\mathcal{B}^{(i)}_{k})$ for every $i$;
      the mask realizes each same-stage $\mathcal{B}^{(i)}_{k}$ from the preceding
      $H$ chunks in this forward.
    \State If $k>k^\star$, set $x^{\mathrm R}_{i,k-1}\gets x^{\mathrm R}_{i,k}
      -(\sigma_k{-}\sigma_{k-1})\widehat v^{\mathrm R}_{i,k}$ for every $i$.
  \EndFor
  \State $x_{0,i}\gets x^{\mathrm R}_{i,k^\star}
    -\sigma_{k^\star}\widehat v^{\mathrm R}_{i,k^\star}$ for every $i$; \Return $(x_0,k^\star)$.
\EndFunction
\Function{TiledScore}{$F$, $x_0$, $x_{\widetilde\sigma}$, $\varepsilon^{\mathrm h}$,
  $c$, $\widetilde\sigma$, $\widetilde\tau$, $g$}
  \State \emph{Tile} $\mathcal{I}$ into
    $(L{-}1)/(W_{\mathrm{score}}{-}1){=}4$ tiles $T_0{=}[0,21)$,
    $T_1{=}[21,41)$, $T_2{=}[41,61)$, $T_3{=}[61,81)$.
    \State $img_{T_0},img_{T_1},img_{T_2}\gets\mathcal{D}(x_0\text{ before }T_3))$;$h_{\ell,0}\gets\mathcal{E}(img_{T_{\ell-1}}),\forall \ell\in\{1,2,3\}$.
  \State $\widehat v[T_0]\gets F(x_{\widetilde\sigma}[T_0],c,\widetilde\tau)$
    at guidance $g$.
  \For{$\ell=1,2,3$}
    \State $h_{\ell,\widetilde\sigma}\gets
      \sg{(1{-}\widetilde\sigma)h_{\ell,0}+\widetilde\sigma\varepsilon^{\mathrm h}}$.
    \State $u_\ell\gets h_{\ell,\widetilde\sigma}\|x_{\widetilde\sigma}[T_\ell]$;
      evaluate $F(u_\ell,c,\widetilde\tau)$ at guidance $g$ and store the prediction in $\widehat v[T_\ell]$.
  \EndFor
  \State \Return $\widehat v$ on $\mathcal{I}$; discard the image-head predictions.
\EndFunction
\For{each training step with caption $c$}
  \If{generator phase}
  \State $(x_0,k^\star)\gets{}$\Call{Rollout}{$c$}.
    \State Sample $\widetilde\sigma$, set
      $\widetilde\tau\gets\tau(\widetilde\sigma)$, and sample
      $\varepsilon,\varepsilon^{\mathrm h}\sim\mathcal{N}(0,I)$;
      $x_{\widetilde\sigma}\gets
      (1{-}\widetilde\sigma)x_0+\widetilde\sigma\varepsilon$.
    \State $\widehat v^{\mathrm{real}}\gets{}$\Call{TiledScore}{$F^{\mathrm{real}},x_0,x_{\widetilde\sigma},\varepsilon^{\mathrm h},c,\widetilde\sigma,\widetilde\tau,3.0$}.
    \State $\widehat v^{\mathrm{fake}}\gets{}$\Call{TiledScore}{$F^{\mathrm{fake}}_\phi,x_0,x_{\widetilde\sigma},\varepsilon^{\mathrm h},c,\widetilde\sigma,\widetilde\tau,1.0$}.
    \State $\widehat x_0^{\mathrm{real}}\gets
      x_{\widetilde\sigma}-\widetilde\sigma\widehat v^{\mathrm{real}}$;
      $\widehat x_0^{\mathrm{fake}}\gets
      x_{\widetilde\sigma}-\widetilde\sigma\widehat v^{\mathrm{fake}}$.
    \State $y\gets\sg{\,x_0-\big(\widehat x_0^{\mathrm{fake}}-\widehat x_0^{\mathrm{real}}\big)\big/
      \mathbb{E}\big|x_0-\widehat x_0^{\mathrm{real}}\big|\,}$ \Comment{distribution-matching target}
    \State $\mathcal{L}_{\mathrm{gen}}\gets\frac{1}{L}\sum_{p\in\mathcal{I}}\tfrac12\|x_{0,p}-y_p\|^2$;
      backpropagate; take a student AdamW step; update the student EMA.
  \Else
  \State $(x_0,k^\star)\gets{}$\Call{Rollout}{$c$}; sample
      $\widetilde\sigma$, set $\widetilde\tau\gets\tau(\widetilde\sigma)$, and
      sample $\varepsilon,\varepsilon^{\mathrm h}\sim\mathcal{N}(0,I)$; form
      $x_{\widetilde\sigma}$ as above.
    \State $\widehat v^{\mathrm{fake}}\gets{}$\Call{TiledScore}{$F^{\mathrm{fake}}_\phi,x_0,x_{\widetilde\sigma},\varepsilon^{\mathrm h},c,\widetilde\sigma,\widetilde\tau,1.0$}.
    \State $\mathcal{L}_\phi\gets\frac{1}{L}\sum_{p\in\mathcal{I}}\|\widehat v^{\mathrm{fake}}_p-(\varepsilon_p-x_{0,p})\|^2$.
    \State Backpropagate $\mathcal{L}_\phi$ through $F^{\mathrm{fake}}_\phi$ only;
      take a critic AdamW step.
  \EndIf
\EndFor
\Ensure The distilled four-step guidance-free student and its EMA model.
\end{algorithmic}
\end{algorithm}
The terminal endpoint is
$\sigma_0=0$.
At each iteration, all ranks sample the same student exit $k^\star$ uniformly from the four stages and retain gradients only at that stage, following the stochastic gradient truncation of Self-Forcing~\citep{huang2025selfforcing}; the higher-noise stages run under stop-gradient. The rollout first generates the three planner blocks and then all $L=81$ renderer positions, stopping at $k^\star$ and using the corresponding $\widehat{x}_0$ as its output. A clean-endpoint cache-extraction forward after each planner block constructs $\mathcal{A}$. We detach this bank when it is read by later planner blocks but keep the renderer's read of $\mathcal{A}_{\Gamma(i)}$ connected at the student exit. Since the renderer produces every output position, this renderer-to-anchor connection is the only path by which the generator loss trains the planner. Alg.~\ref{alg:dmd} gives the complete rollout and loss computation.

At a fixed renderer stage, training evaluates all chunks in one packed forward.
The block-causal renderer mask exposes to chunk $i$ only its own latents, the
anchor window $\mathcal{A}_{\Gamma(i)}$, and at most the $H$ preceding chunks
from that same stage.
Inference materializes the equivalent same-stage bank sequentially through
Eq.~(\ref{eq:step-update}), whereas the packed training forward retains the
same dependencies and permits gradients through their KV edges.

\paragraph{Windowed teacher and critic scores.}
Each score network accepts at most $W_{\mathrm{score}}=21$ latent-frame positions.  Alg.~\ref{alg:dmd} therefore divides the query into four
target tiles while preserving the full-length student rollout and gradient
path. Following LongLive~\citep{yang2026longlive}, we prepend to each later tile a detached image head obtained by re-encoding the last RGB frame of the decoded clean prefix. 
This head predictions are discarded after getting the scores.

\paragraph{Optimization.}
The frozen real-score is Wan2.1-T2V-14B, and the critic is Wan2.1-T2V-1.3B with a trainable rank-$256$ LoRA; the full student remains trainable. We run $600$ AdamW steps of global batch size $128$ at a $1{:}5$ generator/critic ratio ($100$ generator updates). 
Generator and critic-LoRA learning rates are $10^{-5}$ and $10^{-6}$, respectively; EMA decay is $0.999$.
The real-data score and critic use guidance scales of $3.0$ and $1.0$, respectively.

\end{document}